\documentclass[nohyperref]{article}

\usepackage{microtype}
\usepackage{graphicx}
\usepackage{subfigure}
\usepackage{booktabs} 

\usepackage{hyperref}

\usepackage[accepted]{icml2022}

\usepackage{amsmath}
\usepackage{amssymb}
\usepackage{mathtools}
\usepackage{amsthm}

\usepackage[capitalize,noabbrev]{cleveref}

\theoremstyle{plain}

\theoremstyle{definition}

\theoremstyle{remark}

\usepackage{hyperref}

\usepackage[textsize=tiny]{todonotes}

\usepackage{xcolor}

\usepackage{paralist}

\usepackage{bbding}
\usepackage{bm}
\usepackage{multirow}
\usepackage{booktabs}
\usepackage{makecell}
\usepackage{algorithm}
\usepackage{algorithmic}

\newtheorem{thm}{Theorem}
\usepackage{verbatim}

\makeatletter
\DeclareRobustCommand\onedot{\futurelet\@let@token\@onedot}
\def\@onedot{\ifx\@let@token.\else.\null\fi\xspace}

\makeatother

\icmltitlerunning{MAE-DET: Maximum Entropy Principle for Efficient Object Detection}

\begin{document}

\twocolumn[
\icmltitle{MAE-DET: Revisiting Maximum Entropy Principle in Zero-Shot NAS for Efficient Object Detection}



\icmlsetsymbol{equal}{*}

\begin{icmlauthorlist}
	\icmlauthor{Zhenhong Sun}{equal,comp}
	\icmlauthor{Ming Lin}{equal,comp}
	\icmlauthor{Xiuyu Sun}{comp}
	\icmlauthor{Zhiyu Tan}{comp}
	\icmlauthor{Hao Li}{comp}
	\icmlauthor{Rong Jin}{comp}
\end{icmlauthorlist}

\icmlaffiliation{comp}{Alibaba Group}

\icmlcorrespondingauthor{Xiuyu Sun}{xiuyu.sxy@alibaba-inc.com}

\icmlkeywords{NAS, Zero-Shot, Neural Architecture Search, Object Detection, Maximum Entropy Principle}

\vskip 0.3in
]



\printAffiliationsAndNotice{\icmlEqualContribution} 

\begin{abstract}
    In object detection, the detection backbone consumes more than half of the overall inference cost. Recent researches attempt to reduce this cost by optimizing the backbone architecture with the help of Neural Architecture Search (NAS). However, existing NAS methods for object detection require hundreds to thousands of GPU hours of searching, making them impractical in fast-paced research and development. In this work, we propose a novel zero-shot NAS method to address this issue. The proposed method, named MAE-DET, automatically designs efficient detection backbones via the Maximum Entropy Principle without training network parameters, reducing the architecture design cost to nearly zero yet delivering the state-of-the-art (SOTA) performance. Under the hood, MAE-DET maximizes the differential entropy of detection backbones, leading to a better feature extractor for object detection under the same computational budgets. After merely one GPU day of fully automatic design, MAE-DET innovates SOTA detection backbones on multiple detection benchmark datasets with little human intervention. Comparing to ResNet-50 backbone, MAE-DET is $+2.0\%$ better in mAP when using the same amount of FLOPs/parameters, and is $1.54$ times faster on NVIDIA V100 at the same mAP. Code and pre-trained models are available at \href{https://github.com/alibaba/lightweight-neural-architecture-search}{https://github.com/alibaba/lightweight-neural-architecture-search}.
\end{abstract}

\section{Introduction}

Seeking better and faster deep models for object detection is never an outdated task in computer vision. The performance of a deep object detection network heavily depends on the feature extraction backbone~\citep{detnet,detnas}. Currently, most state-of-the-art (SOTA) detection backbones~\citep{resnet,resnext,dcn,li2021ds} are designed manually by human experts, which can take years to develop. Since the detection backbone consumes more than half of the total inference cost in many detection frameworks, it is critical to optimize the backbone architecture for better speed-accuracy trade-off on different hardware platforms, ranging from server-side GPUs to mobile-side chipsets. To reduce time cost and human labor, Neural Architecture Search (NAS) has emerged to facilitate the architecture design. Various NAS methods have demonstrated their efficacy in designing SOTA image classification models~\citep{nasnet,darts,ofa,efficient,fna}. These successful stories inspire recent researchers to use NAS to design detection backbones~\citep{detnas,peng2019efficient,mobiledets,spinenet,spnas} in an end-to-end way.

To date, existing NAS methods for detection task are all training-based, meaning they need to train network parameters to evaluate the performance of network candidates on the target dataset, a process that consumes enormous hardware resources. This makes the training-based NAS methods inefficient in modern fast-paced research and development. To reduce the searching cost, training-free methods are recently proposed, also known as \textit{zero-shot NAS} in some literatures~\citep{syn,naswot,ntk,zennas}. The zero-shot NAS predicts network performance without training network parameters, and is therefore much faster than training-based NAS. As a relatively new technique, existing zero-shot NAS methods are mostly validated on classification tasks. Applying zero-shot NAS to detection task is still an intact challenge.

In this work, we present the first effort to introduce zero-shot NAS technique to design efficient object detection backbones. We show that directly transferring existing zero-shot NAS methods from image classification to detection backbone design will encounter fundamental difficulties. While image classification network only needs to predict the class probability, object detection network needs to additionally predict the bounding boxes of multiple objects, making the direct architecture transfer sub-optimal. To this end, a novel zero-shot NAS method, termed MAximum-Entropy DETection (MAE-DET), is proposed for searching object detection backbones. The key idea behind MAE-DET is inspired by the \textit{Maximum Entropy Principle} ~\citep{jaynes57_InformationTheory,theory1,theory2,theory3}. Informally speaking, when a detection network is formulated as an information processing system, its capacity is maximized when its entropy achieves maximum under the given inference budgets, leading to a better feature extractor for object detection. Based on this observation, MAE-DET maximizes the differential entropy~\citep{shannon1948} of detection backbones by searching for the optimal configuration of network depth and width without training network parameters.

The Principle of Maximum Entropy is one of the fundamental first principles in Physics and Information Theory. As well as the widespread applications of deep learning, many theoretical studies attempt to understand the success of deep learning from the Maximum Entropy Principle~\citep{michael2018OnInformationBottleneckTheory,chanReduNetWhiteboxDeep2021,LearningDiverseNIPS2020}. Inspired by these pioneer works, MAE-DET establishes a connection from Maximum Entropy Principle to zero-shot object detection NAS. This leads to a conceptually simple design, yet endowed with strong empirical performance. Only using the standard single-branch convolutional blocks, the MAE-DET can outperform previous detection backbones built by much more involved engineering. This encouraging result again verifies an old-school doctrine: \textbf{simple is better}.

While the Maximum Entropy Principle has been applied in various scientific problems, its application in zero-shot NAS is new. Particularly, a direct application of the principle to object detection will raise several technical challenges. \textbf{The first challenge is how to estimate the entropy of a deep network}. The exact computation of entropy requires knowing the precise probability distribution of deep features in high dimensional space, which is difficult to estimate in practice. To address this issue, MAE-DET estimates the Gaussian upper bound of the differential entropy, which only requires to estimate the variance of the feature maps. \textbf{The second challenge is how to efficiently extract deep features for objects of different scales}. In real-world object detection datasets, such as MS COCO~\citep{coco}, the distribution of object size is data-dependent and non-uniform. To bring this prior knowledge in backbone design, we introduce the \textit{Multi-Scale Entropy Prior} (MSEP) in the entropy estimation. We find that the MSEP significantly improves detection performance. The overall computation of MAE-DET takes one forward inference of the detection backbone, therefore its cost is nearly zero compared to previous training-based NAS methods.

The contributions of this work are summarized as follows:
\begin{compactitem}
\item[$\bullet$] We revisit the Maximum Entropy Principle in zero-shot object detection NAS. The proposed MAE-DET is conceptually simple, yet delivers superior performance without bells and whistles.
\item[$\bullet$] Using less than one GPU day and 2GB memory, MAE-DET achieves competitive performance over previous NAS methods on COCO with at least 50x times faster.
\item[$\bullet$] MAE-DET is the first zero-shot NAS method for object detection with SOTA performance on multiple benchmark datasets under multiple detection frameworks. 
\end{compactitem}

\section{Related work} 
\noindent\textbf{Backbone Design for Object Detection}$\quad$
Recently, the design of object detection models composing of backbone, neck and head has become increasingly popular due to their effectiveness and high performance~\citep{fpn,retinanet,fcos,gfv1,gfv2,bifpn,giraffedet}.
Prevailing detectors directly use the backbones designed for image classification to extract multi-scale features from input images, such as ResNet~\citep{resnet}, ResNeXt~\citep{resnext} and Deformable Convolutional Network (DCN)~\citep{dcn}.
Nevertheless, the backbone migrated from image classification may be suboptimal in object detection \citep{nasfpn}. 
To tackle the gap, several architectures optimized for object detection are proposed, including Stacked Hourglass~\citep{hourglass}, FishNet~\citep{fishnet}, DetNet~\citep{detnet}, 
HRNet~\citep{hrnet} and so on. Albeit with good performance, these hand-crafted architectures heavily rely on expert knowledge and a tedious trial-and-error design.

\noindent\textbf{Neural Architecture Search}$\quad$ 
Neural Architecture Search (NAS) is initially developed to automatically design network architectures for image classification~\citep{nasnet,darts,abenas,ofa,darts-,mcunet,efficient,autoformer,zennas}. Using NAS to design object detection models has not been well explored. Currently, existing detection NAS methods are all training-based methods. Some methods focus on searching detection backbones, such as DetNAS~\citep{detnas}, SpineNet~\citep{spinenet} and SP-NAS~\citep{spnas}, while others focus on searching FPN neck, such as NAS-FPN~\citep{nasfpn}, NAS-FCOS~\citep{nasfcos} and OPANet~\citep{opanas}. These methods require training and evaluation of the target datasets, which is intensive in computation. MAE-DET distinguishes itself as the first zero-shot NAS method for the backbone design of object detection.

\section{Preliminary}
\label{sec:Preliminary}

In this section, we first formulate a deep network as a system endowed with continuous state space. Then we define the differential entropy of this system and show how to estimate this entropy via its Gaussian upper bound. Finally, we introduce the basic concept of vanilla network search space for designing our detection backbones.

\noindent\textbf{Continuous State Space of Deep Networks}$\quad$ 
A deep network $F(\cdot): \mathbb{R}^{d} \xrightarrow{} \mathbb{R}$ maps an input image $\boldsymbol{x}\in \mathbb{R}^d$ to its label $y \in \mathbb{R}$. The topology of a network can be abstracted as a graph $\mathcal{G}=(\mathcal{V}, \mathcal{E})$ where the vertex set $\mathcal{V}$ consists of neurons and the edge set $\mathcal{E}$ consists of spikes between neurons. For any $v \in \mathcal{V}$ and $e \in \mathcal{E}$, $h(v) \in \mathbb{R}$ and $h(e)\in \mathbb{R}$ present the values endowed with each vertex $v$ and each edge $e$ respectively. The set $\mathcal{S} = \{ h(v),h(e): \forall v\in \mathcal{V}, e \in \mathcal{E}\}$ defines the continuous state space of the network $F$.

According to the Principle of Maximum Entropy, we want to maximize the differential entropy of network $F$, under some given computational budgets. The entropy $H(\mathcal{S})$ of set $\mathcal{S}$ measures the total information contained in the system (network) $F$, including the information contained in the latent features $H(\mathcal{S}_v) = \{h(v): v \in \mathcal{V}\}$ and in the network parameters  $H(\mathcal{S}_e) = \{h(e): e \in \mathcal{E}\}$. As for object detection backbone design, we only care about the entropy of latent features $H(\mathcal{S}_v)$ rather than the entropy of network parameters $H(\mathcal{S}_e)$. Informally speaking, $H(\mathcal{S}_v)$ measures the feature representation power of $F$ while $H(\mathcal{S}_e)$ measures the model complexity of $F$.  Therefore, in the remainder of this work, the differential entropy of $F$ refers to the entropy $H(\mathcal{S}_v)$ by default.

\noindent\textbf{Entropy of Gaussian Distribution}$\quad$
The differential entropy of Gaussian distribution can be found in many textbooks, such as \citep{entropy}.
Suppose $x$ is sampled from Gaussian distribution $\mathcal{N}(\mu,\sigma^2)$. Then the differential entropy of $x$ is given by
\begin{equation}
	\label{eq:entrop}
H^*(x) =\frac{1}{2}\log(2\pi) + \frac{1}{2} + H(x) \quad H(x):=\log(\sigma)\,.
\end{equation}
From Eq.~\ref{eq:entrop}, the entropy of Gaussian distribution only depends on the variance. In the following, we use $H(x)$ instead of $H^*(x)$ as constants do not matter in our discussion.

\noindent\textbf{Gaussian Entropy Upper Bound} Since the probability distribution $\mathbb{P}(\mathcal{S}_v)$ is a high dimensional function, it is difficult to compute the precise value of its entropy directly. Instead, we propose to estimate the upper bound of the entropy, given by the following well-known theorem~\citep{coverElementsInformationTheory2012}:

\begin{thm}
	\label{thm:gaussian-upper-bound-entropy} For any continuous distribution $\mathbb{P}(x)$ of mean $\mu$ and variance $\sigma^2$, its differential entropy is maximized when $\mathbb{P}(x)$ is a Gaussian distribution $\mathcal{N}(\mu,\sigma^2)$.
\end{thm}

Theorem \ref{thm:gaussian-upper-bound-entropy} says the differential entropy of a distribution is upper bounded by a Gaussian distribution with the same mean and variance. Combining this with Eq. (\ref{eq:entrop}), we can easily estimate the network entropy $H(\mathcal{S}_v)$ by simply computing the feature map variance and then using Eq. (\ref{eq:entrop}) to get the Gaussian entropy upper bound for the network.

\noindent\textbf{Vanilla Network Search Space} Following previous works, we design our backbones in the vanilla convolutional network space~\citep{detnet,detnas,spinenet,zennas}. This space is one of the most simple spaces proposed in the early age of deep learning, and is now widely adopted in detection backbones. It is also a popular prototype in many theoretical studies~\citep{exponential,bounding,complexity}.

A vanilla network is stacked by multiple convolutional layers, followed by RELU activations. Consider a vanilla convolutional network with $D$ layers of weights ${\bf {W}}^1$,  $...$,  ${\bf {W}}^D$ whose output feature maps are $\ {\boldsymbol {x}}^1$, $...$, ${\boldsymbol {x}}^D$. The input image is $\boldsymbol{x}^0$. Let $\phi(\cdot)$ denote the RELU activation function. Then the forward inference is given by
\begin{align}
	{\boldsymbol {x}}^{l} =  \phi({\boldsymbol {h}}^{l}), {\boldsymbol {h}}^{l}={{\bf {W}}^l} * {\boldsymbol {x}}^{l-1} \quad l=1, ..., D \, .
	\label{eq:forward}
\end{align}
For simplicity, we set the bias of the convolutional layer to zero.

\noindent\textbf{Simple is Better}$\quad$ The vanilla convolutional network is very simple to implement. Most deep learning frameworks~\cite{Pytorch_NIPS2019,tensorflow2015-whitepaper} provide well optimized convolutional operators on GPU. The training of convolutional networks is well studied, such as adding residual link~\citep{resnet} and Batch Normalization (BN)~\citep{bn} will greatly improve convergence speed and stability. While we stick to the simple vanilla design on purpose, the building blocks used in MAE-DET can be combined with other auxiliary components to ``modernize'' the backbone to boost performance, such as Squeeze-and-Excitation (SE) block~\citep{hu2018squeeze} or self-attention block~\citep{zhao2020exploring}. Thanks to the simplicity of MAE-DET, these auxiliary components can be easily plugged into the backbone without special modification. Once again, we deliberately avoid using these auxiliary components to keep our design simple and universal. By default, we only use residual link and BN layer to accelerate the convergence. In this way, it is clear that the improvements of MAE-DET indeed come from a better backbone design.

\section{Maximum Entropy Zero-Shot NAS for Object Detection}

In this section, we first describe how to compute the differential entropy for deep networks. Then we introduce the Multi-Scale Entropy Prior (MSEP) to better capture the prior distribution of object size in real-world images. Finally, we present the complete MAE-DET backbone designed by a customized Evolutionary Algorithm (EA).

\begin{figure}[htb]
	\centering
	\includegraphics[scale=0.22]{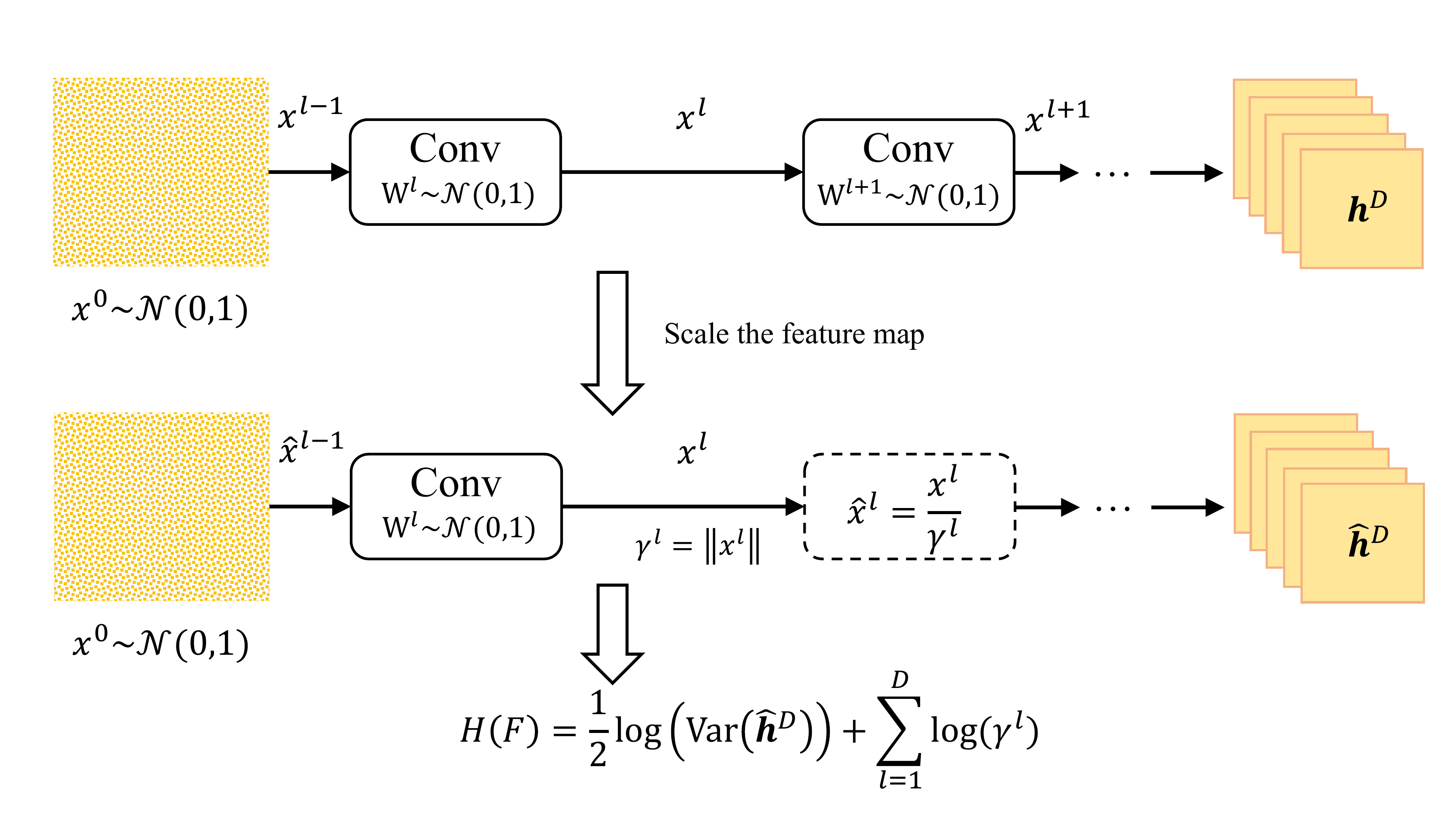}
	\caption{Single-scale entropy with rescaling for deep networks.}
	\label{fig:escore}	
\end{figure}

\subsection{Differential Entropy for Deep Networks}
In this subsection, we present the computation of differential entropy for the final feature map generated by a deep network. First, all parameters are initialized by the standard Gaussian distribution $\mathcal{N}(0,1)$. Then we randomly generate an image filled with the standard Gaussian noise and perform forward inference. Based on the discussion in Section \ref{sec:Preliminary}, the (Gaussian upper bound) entropy $H(F)$ of the network $F$ is given by
\begin{align}
	\label{eq:entropy-of-unscaled-network}
	H(F) = \frac{1}{2} \log( \mathrm{Var}( \boldsymbol{h}^D)) \ . 
\end{align}
Please note the variance is computed on the last pre-activation feature map $\boldsymbol{h}^D$.

For deep vanilla networks, directly using Eq. (\ref{eq:entropy-of-unscaled-network}) might cause numerical overflow. This is because every layer amplifies the output norm by a large factor. The same issue is also reported in Zen-NAS~\citep{zennas}. Inspired by the BN rescaling technique proposed in Zen-NAS, we propose an alternative solution without BN layers. We directly re-scale each feature map $\boldsymbol{x}^l$ by some constants $\gamma^l$ during inference, that is ${\boldsymbol {x}}^{l} =  \phi({\boldsymbol {h}}^{l}) / \gamma^l$, and then compensate the entropy of the network by
\begin{align}
	\label{eq:entropy-of-rescaled-network}
	H(F) = \frac{1}{2} \log( \mathrm{Var}( \boldsymbol{\hat h}^D)) + \sum_{l=1}^{D} \log(\gamma^l) \ . 
\end{align}
 
The values of $\gamma^l$ can be arbitrarily given, as long as the forward inference does not overflow or underflow. In practice, we find that simply setting $\gamma^l$ to the Euclidean norm of the feature map works well. The process is illustrated in Figure \ref{fig:escore}. Finally, $H(F)$ is multiplied by the size of the feature map as the entropy estimation for this feature map.

\noindent\textbf{Compare with Zen-NAS}$\quad$ The principles behind MAE-DET and Zen-NAS are fundamentally different. 
Zen-NAS uses the gradient norm of the input image as ranking score, and proposes to use two feed-forward inferences to approximate the gradient norm for classification. In contrast, MAE-DET uses an entropy-based score, which only requires one feed-forward inference. Please see the Experiment section for more empirical comparisons.

\begin{figure}[tb]
	\centering
	\includegraphics[scale=0.4]{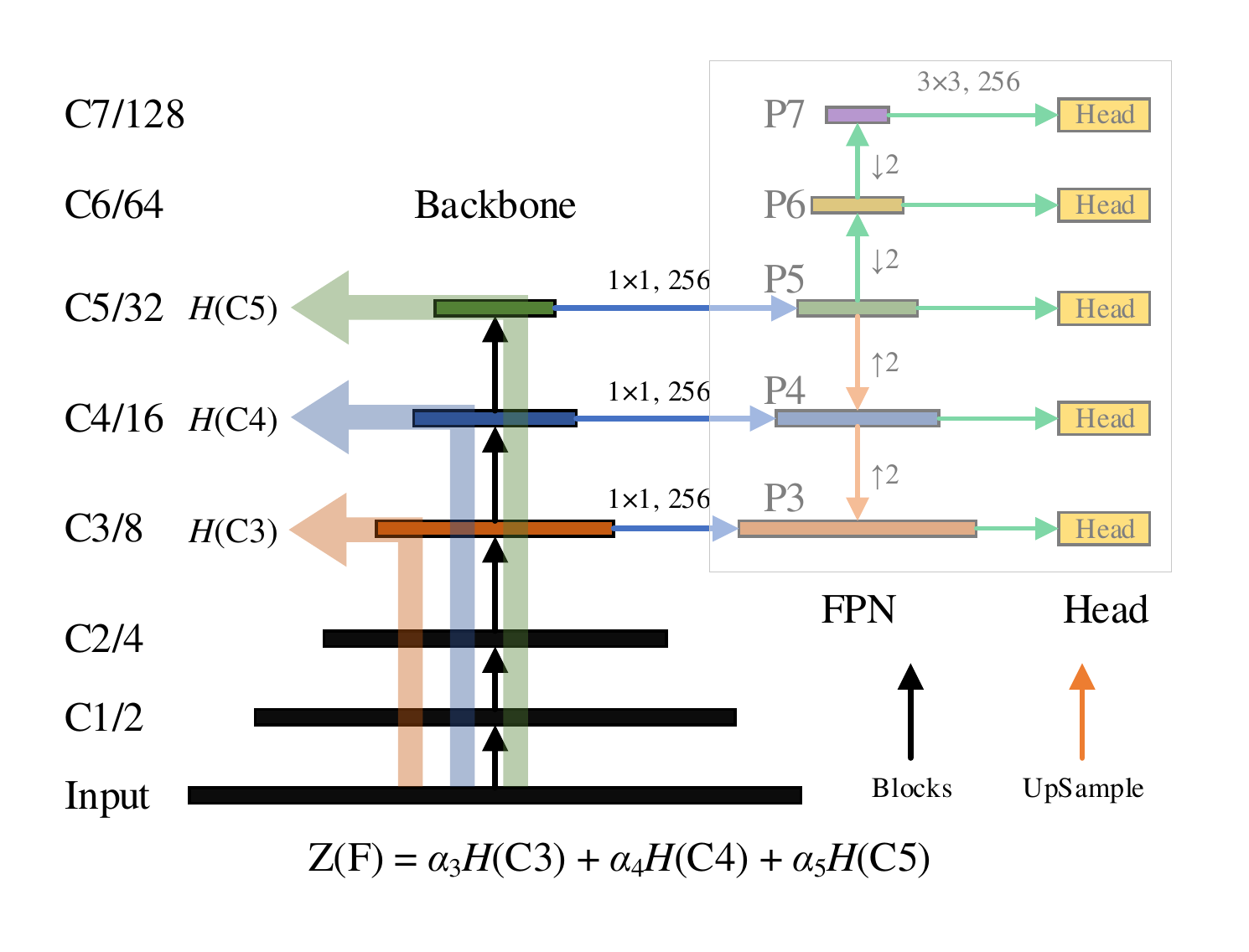}
	\caption{Multi-scale entropy for detection backbone with multi-scale features.}
	\label{fig:mescore}	
\end{figure}
\subsection{Multi-Scale Entropy Prior (MSEP) for Object Detection}\label{sub:msep}
In real-world images, the distribution of object size is not uniform. To bring in this prior knowledge, the detection backbone has 5 stages, where each stage downsamples the feature resolution to half. The MSEP collects the feature map from the last layer of each stage and weighted-sum the corresponding feature map entropies as a new measurement. We name this new measurement multi-scale entropy. The process is illustrated in Figure \ref{fig:mescore}. In this figure, the backbone extracts multi-scale features $\boldsymbol{C} = ({\rm C}1, {\rm C}2, ..., {\rm C}5)$ at different resolutions. Then the FPN neck fuses $\boldsymbol{C}$ as input features $\boldsymbol{P} = ({\rm P}1, {\rm P}2, ..., {\rm P}7)$ for the detection head. The multi-scale entropy $Z(F)$ of backbone $F$ is then defined by
\begin{align}
	\label{eq:MSEP-score}
	{Z(F)} := \alpha_1 H({\rm C}1) + \alpha_2 H({\rm C}2) + \cdots + \alpha_5 H({\rm C}5)
\end{align}
where $H(Ci)$ is the entropy of $Ci$ for $i=1,2,\cdots,5$. The weights $\boldsymbol{\alpha}=(\alpha_1, \alpha_2, \cdots, \alpha_5)$ store the multi-scale entropy prior to balance the expressivity of different scale features. 

\noindent \textbf{How to choose ${\alpha}$}$\quad$
As a concrete example in Fig.~\ref{fig:mescore}, the parts of ${\rm P}3$ and ${\rm P}4$ are generated by up-sampling of ${\rm P}5$, and ${\rm P}6$ and ${\rm P}7$ are directly generated by down-sampling of ${\rm P}5$ (generated by ${\rm C}5$).
Meanwhile, based on the fact that ${\rm C}5$ carries sufficient context for detecting objects on various scales~\citep{yolof}, ${\rm C}5$ is important in the backbone search, so it is good to set a larger value for the weight $ \alpha_{5}$.
Then, different combinations of ${\alpha}$ and correlation analysis are explored in Appendix~\ref{app:weight}, indicating that $\boldsymbol{\alpha}=(0,0,1,1,6)$ is good enough for the FPN structure.

\begin{algorithm}
	\caption{MAE-DET with Coarse-to-Fine Evolution}
	\label{alg:MAE-DET}
	\begin{algorithmic}[1]
		
		\REQUIRE Search space $\mathcal{S}$, inference budget $B$, maximal depth $L$, total number of iterations $T$, evolutionary population size $N$, initial structure $F_0$, fine-search flag \textit{Flag}.
		
		\ENSURE NAS-designed MAE-DET backbone $F^*$.
		
		\STATE Initialize population $\mathcal{P}=\{F_0\}$, \textit{Flag=False}.
		
		\FOR{$t=1,2,\cdots,T$}
		\IF{$t$ equals to $T/2$}
		\STATE Keep top 10 networks of highest multi-scale entropy in $\mathcal{P}$ and remove the others.
		\STATE Set $Flag=True$.
		\ENDIF
		\STATE Randomly select $F_t \in \mathcal{P}$.
		\STATE Mutate $\hat{F}_t=\textrm{MUTATE}(F_t, \mathcal{S}, Flag)$
		\IF{$\hat{F}_t$ exceeds inference budget or has more than $L$ layers} 
		\STATE Do nothing.
		\ELSE
		\STATE Get multi-scale entropy $Z(\hat{F}_t)$.
		\STATE Append $\hat{F}_t$ to $\mathcal{P}$.
		\ENDIF
		
		\STATE Remove networks of the smallest multi-scale entropy if the size of $\mathcal{P}$ exceeds $B$.
		\ENDFOR
		
		\STATE Return $F^*$, the network of the highest multi-scale entropy in $\mathcal{P}$.
		
	\end{algorithmic}
\end{algorithm}

\subsection{Evolutionary Algorithm for MAE-DET}

\begin{figure}[H]
	\centering
	\vspace{-0.5cm}
	\includegraphics[scale=0.3]{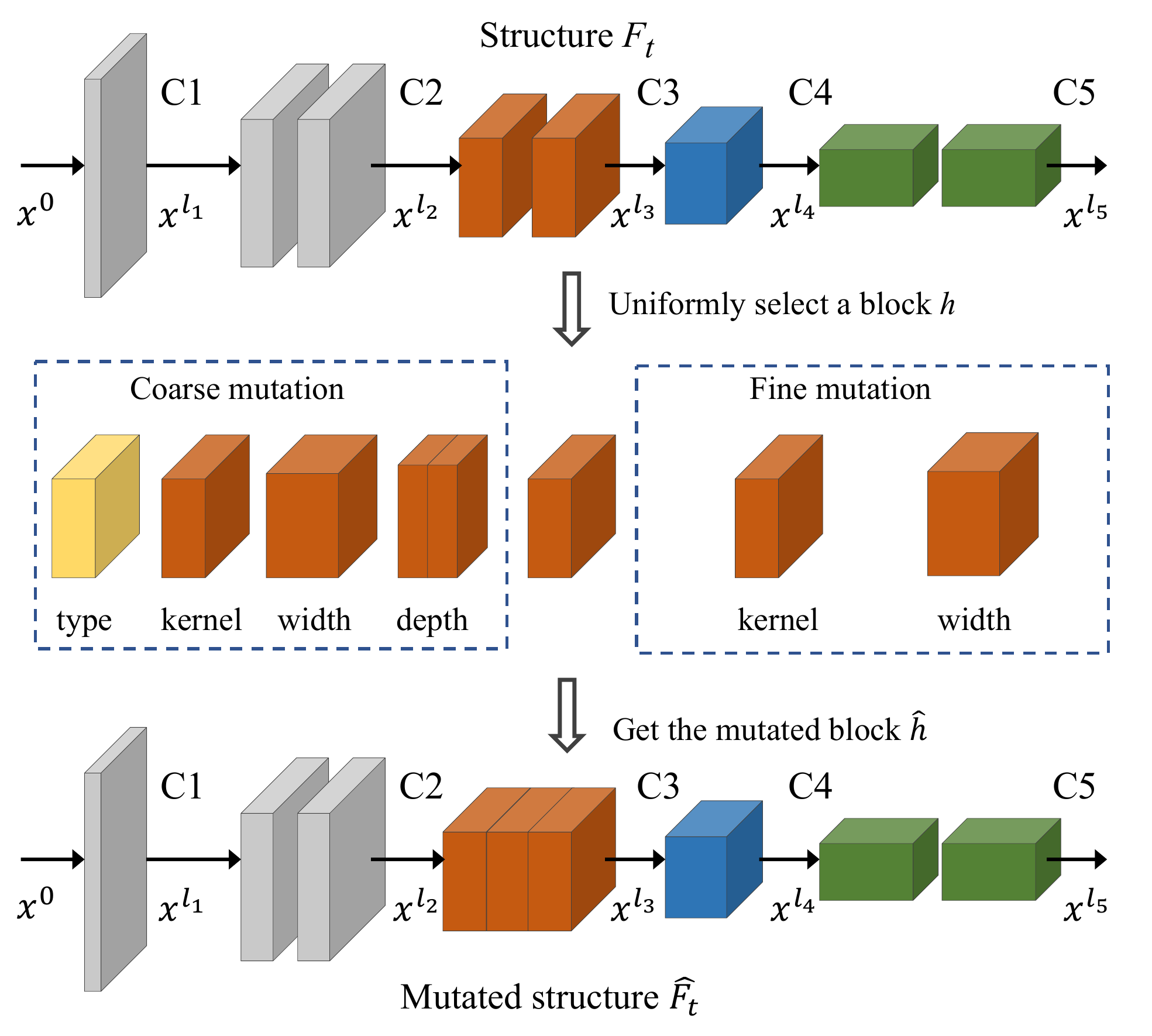}			
	\caption{Visualization of Algorithm~\ref{alg:mutate}.}
	\label{fig:block}			
\end{figure}

Combining all above, we present our NAS algorithm for MAE-DET in Algorithm~\ref{alg:MAE-DET}. The MAE-DET maximizes the multi-scale differential entropy of detection backbones using a customized Evolutionary Algorithm (EA). To improve evolution efficiency, a coarse-to-fine strategy is proposed to reduce the search space gradually. First, we randomly generate $N$ seed architectures to fill the population $\mathcal{P}$. As shown in Figure~\ref{fig:block}, a seed architecture $F_t$ consists of a sequence of building blocks, such as ResNet block~\citep{resnet} or MobileNet block~\citep{mbv2}. Then we randomly select one block and replace it with its mutated version. We use coarse-mutation in the early stages of EA, and switch to fine-mutation after $T/2$ EA iterations.  In the coarse-mutation, the block type, kernel size, depth and width are randomly mutated. In the fine-mutation, only kernel size and width are mutated.

After the mutation, if the inference cost of the new structure $\hat{F}_t$ does not exceed the budget (e.g., FLOPs, parameters and latency) and its depth is smaller than budget $L$, $\hat{F}_t$ is appended into the population $\mathcal{P}$. The maximal depth $L$ prevents the algorithm from generating over-deep structures, which will have high entropy with unreasonable structure, and the performance will be worse. During EA iterations, the population is maintained to a certain size by discarding the worst candidate of the smallest multi-scale entropy. At the end of evolution, the backbone with the highest multi-scale entropy is returned.

\begin{algorithm}[H]
	\caption{MUTATE}
	\label{alg:mutate}
	\begin{algorithmic}[1]
		\REQUIRE Structure $F_t$, search space $\mathcal{S}$, fine-search flag \textit{Flag}. 
		\ENSURE Randomly mutated structure $\hat{F}_t$.
		\STATE Uniformly select a block $h$ in $F_t$.
		\IF{\textit{Flag} equals to \textit{True}}
		\STATE Uniformly alternate the kernel size, width within some range.
		\ELSE
		\STATE Uniformly alternate the block type, kernel size, width and depth within some range.
		\ENDIF
		\STATE Return the mutated structure $\hat{F}_t$.
	\end{algorithmic}
\end{algorithm}

\section{Experiments}

In this section, we first describe detail settings for search and training. Then in Subsection \ref{sec:compare-MAE-DET-vs-resnet}, we apply MAE-DET to design better ResNet-like backbones on COCO dataset~\citep{coco}. We align the inference budget with ResNet-50/101. The performance of MAE-DET and ResNet are compared under multiple detection frameworks, including RetinaNet~\citep{retinanet}, FCOS~\citep{fcos}, and GFLV2~\citep{gfv2}. For fairness, we use the same training setting in all experiments for all backbones. In Subsection \ref{sub:efficiency}, we compare the search cost of MAE-DET to SOTA NAS methods for object detection. 
Subsection \ref{sub:ablation} reports the ablation studies of different components in MAE-DET. Finally, Subsection \ref{sub:transfer} verifies the transferability of MAE-DET on several detection datasets and segmentation tasks.
\textcolor{black}{Due to space limitations, more experiments are postponed to Appendix.} Appendix~\ref{app:mbv2} reports the performance of MAE-DET on mobile devices. Appendix~\ref{app:zero-shot} compare MAE-DET against previous zero-shot NAS methods designed for image classification tasks, showing that these zero-shot NAS methods perform sub-optimally in object detection.

\begin{table*}[th]
	\caption{MAE-DET and ResNet on the COCO. All results using the same training setting. FPS on V100 is benchmarked on the full model with NVIDIA V100 GPU, pytorch, FP32, batch size 32.}
	\label{table:sota}
	\begin{center}
		\scalebox{0.82}{
			\begin{tabular}{ccc|c|cccc|c|cc}
				\toprule[1.5pt]
					& FLOPs & Params & & \multicolumn{4}{c|}{val2017} & test-dev & \multicolumn{1}{c}{FPS}\\
				Backbone & {Backbone} & {Backbone} & Head & ${\bf AP_{val}}$& ${\bf AP}_{S}$ & ${\bf AP}_{M}$ & ${\bf AP}_{L}$ & ${\bf AP_{test}}$ & \makecell[c]{on V100}\\
				\midrule[1.5pt]
				\multirow{3}*{ResNet-50} & \multirow{3}*{83.6G} & \multirow{3}*{23.5M} &
					RetinaNet & 40.2 & 24.3 & 43.3 & 52.2 & - & 23.2 &  \\
					&  &  & FCOS & 42.7 & 28.8 & 46.2 & 53.8 & - & 27.6 &  \\
					&  &  & GFLV2 & 44.7 & 29.1 & 48.1 & 56.6 & 45.1 & 24.2 &  \\
					\midrule[1pt]
				\multirow{3}*{ResNet-101} & \multirow{3}*{159.5G} & \multirow{3}*{42.4M} &
					RetinaNet & 42.1 & 25.8 & 45.7 & 54.1 & - & 18.7 &  \\
					&  &  & FCOS & 44.4 & 28.3 & 47.9 & 56.9 & - & 21.6 &  \\
					&  &  & GFLV2 & 46.3 & 29.9 & 50.1 & 58.7 & 46.5 & 19.4 &  \\
				\midrule[1.5pt]
				\multirow{3}*{MAE-DET-S} & \multirow{3}*{48.7G} & \multirow{3}*{21.2M} &
					RetinaNet & 40.0 & 23.9 & 43.3 & 52.7 & - & 35.5 &  \\
					&  &  & FCOS & 42.5 & 26.8 & 46.0 & 54.6 & - & 43.0 &  \\
					&  &  & GFLV2 & 44.7 & 27.6 & 48.4 & 58.2 & 44.8 & 37.2 &  \\
					\midrule[1pt]
				\multirow{3}*{MAE-DET-M} & \multirow{3}*{89.9G} & \multirow{3}*{25.8M} &
					RetinaNet & 42.0 & 26.7 & 45.2 & 55.1 & - & 21.5 &  \\
					&  &  & FCOS & 44.5 & 28.6 & 48.1 & 56.1 & - & 24.2 &  \\
					&  &  & GFLV2 & 46.8 & 29.9 & 50.4 & 60.0 & 46.7 & 22.2 &  \\
					\midrule[1pt]
				\multirow{3}*{MAE-DET-L} & \multirow{3}*{152.9G} & \multirow{3}*{43.9M} &
					RetinaNet & 43.0 & 27.3 & 46.5 & 56.0 & - & 17.6 &  \\
					&  &  & FCOS & 45.9 & 30.2 & 49.4 & 58.4 & - & 19.2 &  \\
					&  &  & GFLV2 & 47.6 & 30.2 & 51.8 & 60.8 & 48.0 & 18.1 &  \\	
					\bottomrule[1.5pt]
		\end{tabular}}
	\end{center}
\end{table*}

\begin{figure*}[h]
	\centering	
	\subfigure[mAP vs. FLOPs]{\includegraphics[width=0.38\linewidth]{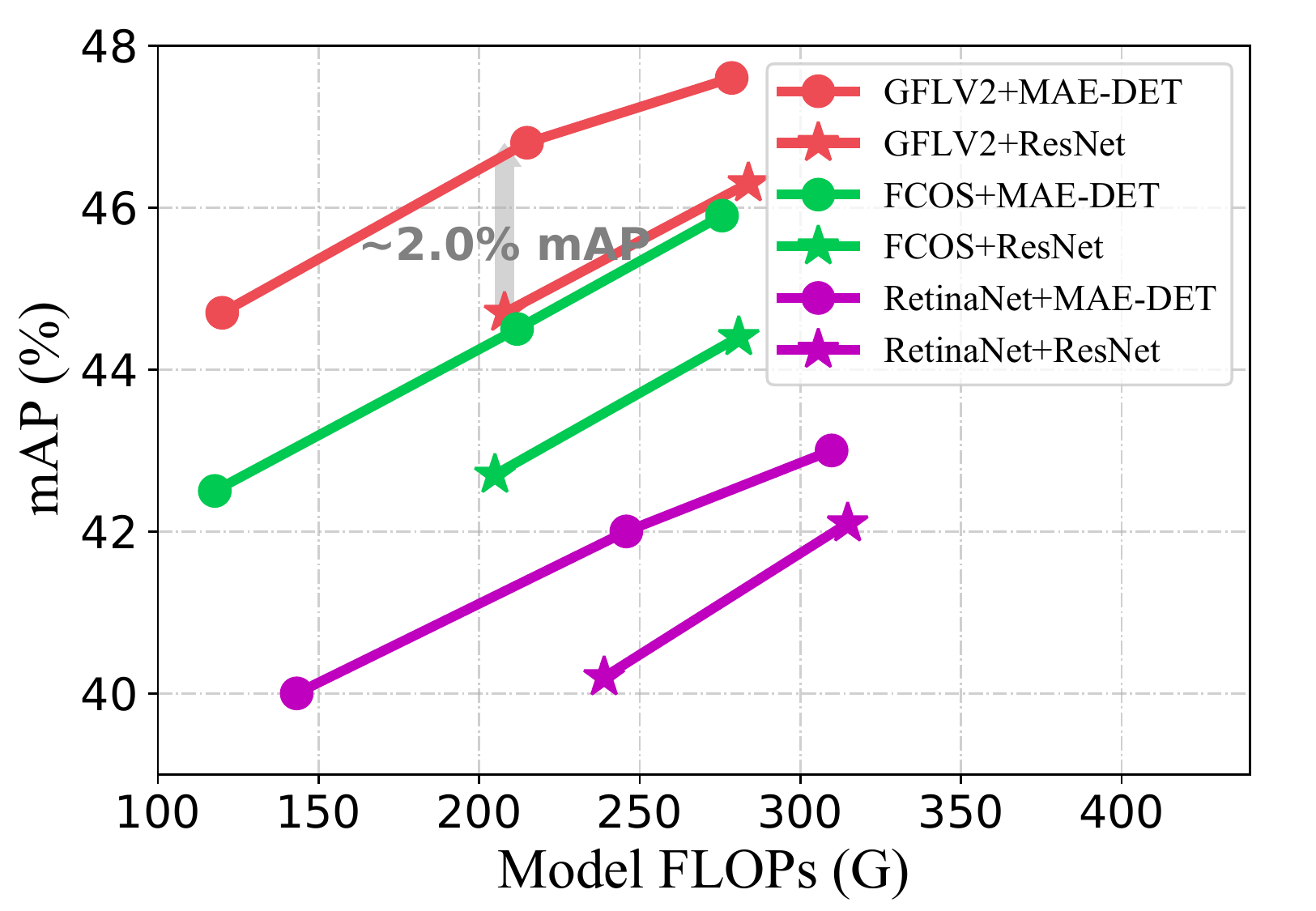} \label{sfig:flops}}
	\subfigure[mAP vs. Speed]{\includegraphics[width=0.38\linewidth]{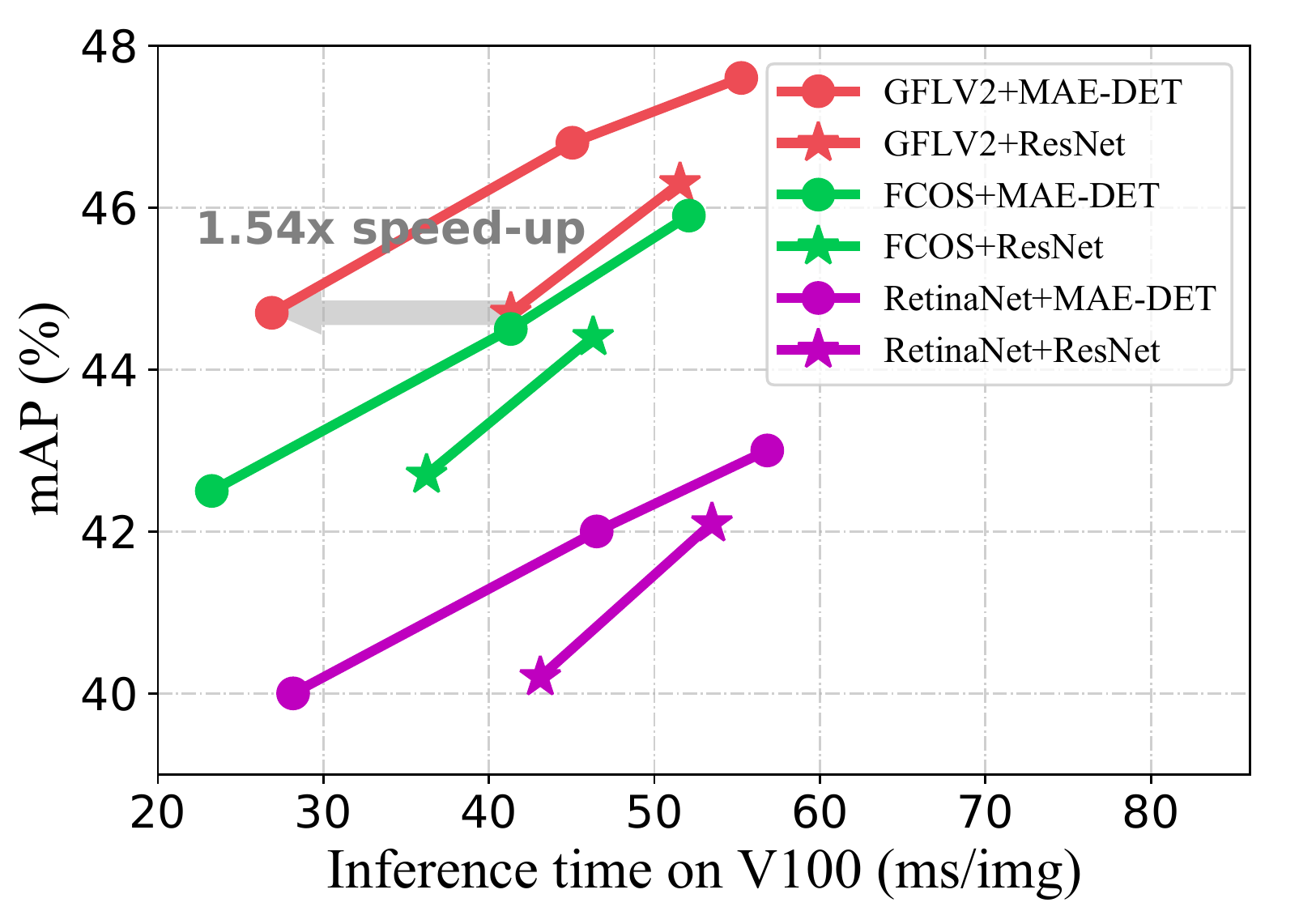} \label{sfig:speed}}	
	\caption{mAP vs. FLOPs and inference speed on COCO val 2017 in Table~\ref{table:sota}. Note that FLOPs in (a) is the value of the full detector, containing backbone, FPN and head.}
	\label{fig:sota}	
\end{figure*}

\subsection{Experiment Settings}
\noindent\textbf{Search Settings}$\quad$ 
In MAE-DET, the evolutionary population $N$ is set to 256. The total EA iterations $T=96000$. 
Following the previous designs~\citep{detnas,spnas,spinenet}, MAE-DET is optimized for FLOPs. The resolution for computing entropy is $384\times384$.

\noindent\textbf{Dataset and Training Details}$\quad$ 
We evaluate detection performance on COCO~\citep{coco} using the official training/testing splits. The mAP is evaluated on val 2017 by default, and GFLV2 is additionally evaluated on test-dev 2007 following common practice. All models are trained from scratch~\citep{scratch} for 6X (73 epochs) on COCO. Following the Spinenet~\citep{spinenet}, we use multi-scale training and Synchronized Batch Normalization (SyncBN). For VOC dataset, train-val 2007 and train-val 2012 are used for training, and test 2007 for evaluation. For image classification, all models are trained on ImageNet-1k~\citep{imagenet} with a batch size of 256 for 120 epochs. Other setting details can be found in Appendix~\ref{app:detail}.

\subsection{Design Better ResNet-like Backbones}
\label{sec:compare-MAE-DET-vs-resnet}

We search efficient MAE-DET backbones for object detection and align with ResNet-50/101 in Table \ref{table:sota}. MAE-DET-S uses 60\% less FLOPs than ResNet-50; MAE-DET-M is aligned with ResNet-50 with similar FLOPs and number of parameters as ResNet-50; MAE-DET-L is aligned with ResNet-101. The feature dimension in the FPN and heads is set to 256 for MAE-DET-M and MAE-DET-L but is set to 192 for MAE-DET-S. The fine-tuned results of models pre-trained on ImageNet-1k are reported in Appendix~\ref{app:pretrain}.

\begin{table*}[h]
	\caption{Comparisons with SOTA NAS methods for object detection. FLOPs are counted for full detector.}
	\label{table:time}
	\begin{center}
		\scalebox{0.8}{
			\begin{tabular}{cccc|ccccc}
				\toprule[1pt]
				Method& \makecell[c]{Training-\\free} &  \makecell[c]{Search Cost\\GPU Days} & Search Part & \makecell[c]{FLOPs\\All} & \makecell[c]{Pretrain/\\Scratch}  & Epochs &  \makecell[c]{COCO \\(${\bf AP_{test}}$)} \\
				\midrule[1pt]
				DetNAS & \bm{$\times$} & $68$ & backbone & 289G &Pretrain & 24 & 43.4\\
				SP-NAS & \bm{$\times$} & $26$ & backbone & 655G & Pretrain & 24 & 47.4\\
				SpineNet & \bm{$\times$} & 100x TPUv3$\dagger$ & backbone+FPN  & 524G & Scratch & 350 & 48.1\\
				\midrule[1pt]
				MAE-DET & \checkmark & \bm{$0.6$} & backbone & 279G & Scratch & 73 & 48.0\\
				\bottomrule[1pt]
		\end{tabular}}		
	\end{center}
	$\dagger$: SpineNet paper did not report the total search cost, only mentioned that 100 TPUv3 was used.
\end{table*}

\begin{table*}[h]
	\caption{Comparisons between MAE-DET, DetNAS~\citep{detnas} and SpineNet~\citep{spinenet} under the same training settings. All backbones are trained under GFLV2 head with 6X training epochs. FLOPs and parameters are counted for full detector.}
	\label{table:fair1}
	\begin{center}
		\scalebox{0.8}{
			\begin{tabular}{ccc|cc|cccc|c}
				\toprule[1pt]
				Backbone & Search Part  & Search Space & \makecell[c]{FLOPs} & \makecell[c]{Params} & ${\bf AP_{val}}$ & ${\bf  AP}_S$ & ${\bf AP}_M$ & ${\bf AP}_L$ & \makecell[c]{FPS\\on V100}\\
				\midrule[1pt]
				DetNAS-3.8G & backbone  & \makecell[c]{ShuffleNetV2\\+Xception} & 205G & 35.5M & 46.4 & 29.3& 50.0 & 59.0 & 17.6 \\
				\midrule[1pt]
				SpineNet-96 & backbone+FPN  & ResNet Block  & 216G & 41.3M & 46.6 & 29.8 & 50.2 & 58.9 & 19.9 \\
				\midrule[1pt]
				MAE-DET-M & backbone  & ResNet Block & 215G  & 34.9M & 46.8 & 29.9 & 50.4 & 60.0 & 22.2  \\
				\bottomrule[1pt]
		\end{tabular}}
	\end{center}	
\end{table*}

In Table \ref{table:sota}, MAE-DET outperforms ResNet by a large margin. The improvements are consistent across three detection frameworks. Particularly, when using the newest framework GFLV2, MAE-DET improves COCO mAP by $+2\%$ at the similar FLOPs of ResNet-50, and speeds up the inference by 1.54x times faster at the same accuracy as ResNet-50. Figure~\ref{fig:sota} visualizes the comparison in Table~\ref{table:sota}.

\noindent \textbf{Remark}$\quad$Please note that we did not copy the numbers of baseline methods reported in previous works in Table~\ref{table:sota} because the mAP depends not only on the architecture, but also on the training schedule, such as training epochs, learning rate, pre-training, etc. Therefore, for a fair comparison, all models in Table~\ref{table:sota} are trained by the same training schedule. For comparison with numbers reported in previous works, see Subsection \ref{sub:efficiency}.

\begin{figure*}[h]
	\centering
	\includegraphics[width=0.8\linewidth]{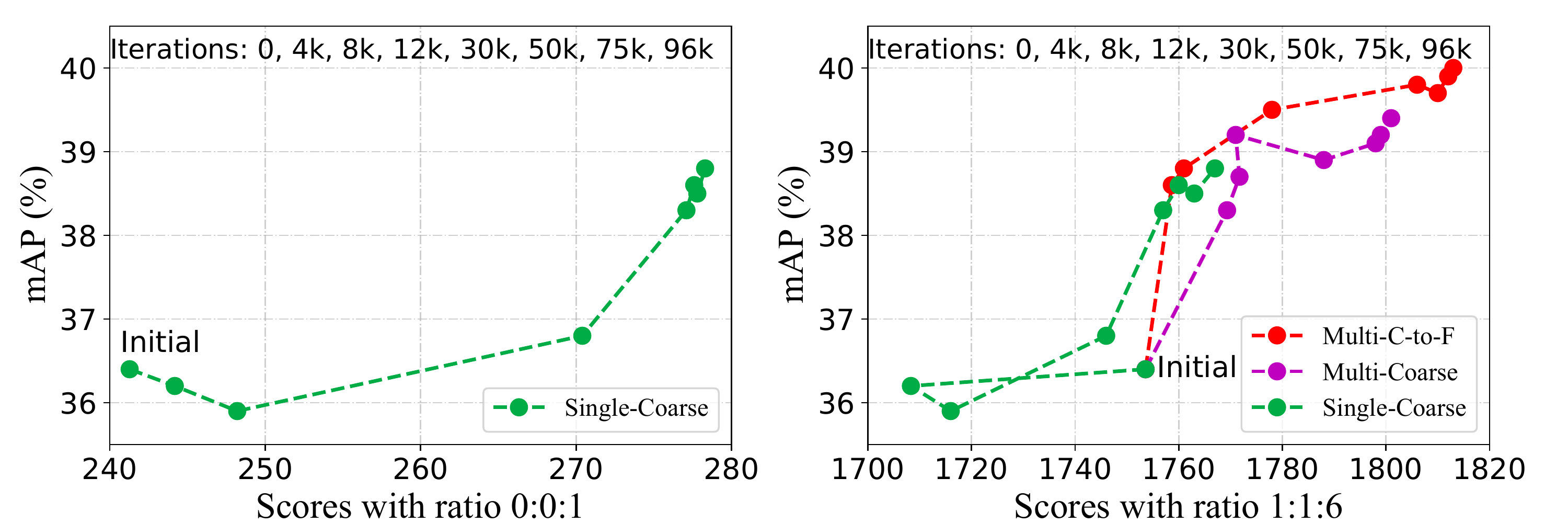}
	\caption{\textcolor{black}{mAP (on FCOS) vs. entropy (scores) during the search with different search strategies. The scores (x-axis) on the left and the right are computed with the ratio of 0:0:1 and 1:1:6 respectively. \textbf{Starting from the initial point, the dotted line indicates the evolution direction in the search process}.}}
	\label{fig:corre}
\end{figure*}

\begin{table*}[h]
	\caption{Comparison of different evolutionary searching strategies in MAE-DET. 
		C-to-F: Coarse-to-Fine. Zen-Score is the proxy in Zen-NAS~\citep{zennas}.}
	\label{table:search}
	\begin{center}
		\scalebox{0.85}{
			\begin{tabular}{cc|ccc|cccc|cccc}
				\toprule[1pt]
				& & \multicolumn{3}{c|}{ImageNet-1K} & \multicolumn{4}{c|}{COCO with YOLOF} & \multicolumn{4}{c}{COCO with FCOS} \\
				Score & Mutation & \makecell[c]{FLOPs} & \makecell[c]{Params} & TOP-1 \% 
				& ${\bf AP_{val}}$ & ${\bf AP}_{S}$ & ${\bf AP}_{M}$ & ${\bf AP}_{L}$ 
				& ${\bf AP_{val}}$ & ${\bf AP}_{S}$ & ${\bf AP}_{M}$ & ${\bf AP}_{L}$\\
				\toprule[1pt]
				ResNet-50 & None & 4.1G & 23.5M & 78.0 & 37.8 & 19.1 & 42.1 & 53.3 & 38.0 & 23.2 & 40.8 & 47.6\\
				\toprule[1pt]
				\textcolor{black}{Zen-Score} & \textcolor{black}{Coarse} & \textcolor{black}{4.4G} & \textcolor{black}{67.9M} 
				& \textcolor{black}{78.9} & \textcolor{black}{38.9} & \textcolor{black}{19.0} & \textcolor{black}{43.2} & \textcolor{black}{56.0}  
				& \textcolor{black}{38.1} & \textcolor{black}{23.2} & \textcolor{black}{40.5} & \textcolor{black}{48.1}\\
				Single-scale  & Coarse & 4.4G & \textbf{60.1M} & 78.7 & 39.8 & 19.9 & 44.4 & 56.5 & 38.8 & 23.1 & 41.4 & 50.1 \\
				Multi-scale & Coarse & 4.3G & 29.4M & 78.9 & 40.1 & 21.1 & 44.5 & 55.9 & 39.4 & 23.7 & 42.3 & 50.0 \\
				Multi-scale & C-to-F & 4.4G & 25.8M & \textbf{79.1} & \textbf{40.3} & \textbf{20.8} & \textbf{44.7} & \textbf{56.4} & \textbf{40.0} & \textbf{24.5} & \textbf{42.6} & \textbf{50.6}\\
				\toprule[1pt]
		\end{tabular}}
	\end{center}	
\end{table*}

\begin{table*}[h]	
	\caption{Transferability of MAE-DET in multiple object detection and instance segmentation tasks. FLOPs reported are counted for full detector.}
	\label{table:mask}
	\begin{center}
		\scalebox{0.8}{
			\begin{tabular}{c|c|c|cccc|cc}
				\toprule[1.5pt]
				Task& Dataset  & Head & Backbone & Resolution & Epochs & \makecell[c]{FLOPs} & ${\bf AP_{val}}$ & ${\bf AP^{mask}_{val}}$\\
				\toprule[1.5pt]
				\multirow{4}*{\makecell[c]{Object\\Detection}} & \multirow{2}*{VOC} & \multirow{4}*{FCOS}
				& ResNet-50 & $1000\times600$ & 12 & 120G & 76.8 & - \\
				& & & MAE-DET-M & $1000\times600$ & 12 & 123G & \textbf{80.9} & - \\
				\cline{2-2}\cline{4-9}
				& \multirow{2}*{Citescapes} & & ResNet-50 & $2048\times1024$ & 64 & 411G & 37.0 & - \\
				& & & MAE-DET-M & $2048\times1024$ & 64 & 426G & \textbf{38.1} & - \\				
				\toprule[1.5pt]
				\multirow{6}*{\makecell[c]{Instance\\Segmentation}} & \multirow{6}*{COCO} & \multirow{4}*{\makecell[c]{MASK\\R-CNN}}
				& ResNet-50 & $1333\times800$ & 73 & 375G & 43.2 & 39.2 \\
				& & & MAE-DET-M & $1333\times800$ & 73  & 379G & \textbf{44.5} & \textbf{40.3} \\
				\cline{4-9}
				& & & ResNet-50$\dagger$ & $640\times640$ & 350  & 228G & 42.7 & 37.8 \\
				& & & SpineNet-49$\dagger$ & $640\times640$ & 350  & 216G & 42.9 & 38.1 \\
				\cline{3-9}
				&  & \multirow{2}*{\makecell[c]{\textcolor{black}{SCNet}}} & \textcolor{black}{ResNet-50} & \textcolor{black}{$1333\times800$} 
				& \textcolor{black}{73} & \textcolor{black}{671G} & \textcolor{black}{46.3} & \textcolor{black}{41.6} \\
				& & & \textcolor{black}{MAE-DET-M} & \textcolor{black}{$1333\times800$} 
				& \textcolor{black}{73}  & \textcolor{black}{675G} & \textcolor{black}{\textbf{47.1}} & \textcolor{black}{\textbf{42.3}} \\
				\toprule[1.5pt]
		\end{tabular}}
	\end{center}
	$\dagger$: Numbers are cited from SpineNet paper~\citep{spinenet}.
\end{table*}

\subsection{Comparison with SOTA NAS Methods}\label{sub:efficiency}

In Table~\ref{table:time}, we compare MAE-DET with SOTA NAS methods for the backbone design in object detection. We directly use the numbers reported in the original papers. Since each NAS method uses different design spaces and training settings, it is impossible to make an absolutely fair comparison for all methods that everyone agrees with. Nevertheless, we list the total search cost, mAP and FLOPs of the best models reported in each work. This gives us an overall impression of how each NAS method works in real-world practice. From Table~\ref{table:time}, MAE-DET is the only zero-shot (training-free) method with $48.0\%$ mAP on COCO, using $0.6$ GPU days of search. SpineNet~\citep{spinenet} achieves a slightly better mAP with 2x more FLOPs. It uses 100 TPUv3 for searching, but the total search cost is not reported in the original paper. MAE-DET achieves better mAP than DetNAS~\citep{detnas} and SP-NAS~\citep{spnas} while being $50\sim 100$ times faster in search.

To further fairly compare different backbones under the same training settings, we train backbones designed by MAE-DET and previous backbone NAS methods in Table~\ref{table:fair1}. Because the implementation of SP-NAS is not open-sourced, we retrain MAE-DET, DetNAS and SpineNet on COCO from scratch. Table~\ref{table:fair1} shows that MAE-DET requires fewer parameters and has a faster inference speed on V100 when achieving competitive performance over DetNAS and SpineNet on COCO.

\subsection{Ablation Study and Analysis}
\label{sub:ablation}

Table~\ref{table:search} reports the MAE-DET backbones searched by different evolutionary strategies, and whether using multi-scale entropy prior. The COCO mAPs of models trained in two detection frameworks (YOLOF and FCOS) are reported in the right big two columns. YOLOF models are trained for 12 epochs with ImageNet pre-trained initialization, while FCOS models are trained with the 3X training epochs. We also compare their image classification ability on ImageNet-1k. All models are constrained by the FLOPs less than 4.4 G while the number of parameters is not constrained. \textcolor{black}{More details about the searching process and architectures can be found in Appendix~\ref{app:vsp},~\ref{app:struc}.}

\noindent\textcolor{black}{\textbf{Single-scale entropy}$\quad$ 
Compared to ResNet-50, the model searched by single-scale entropy obtains $+0.7\%$ accuracy gain on ImageNet, $+2\%$ mAP gain with FPN-free YOLOF and $+0.8\%$ mAP gain with FPN-based FCOS. Meanwhile, the model searched by Zen-Score achieves $+0.9\%$ accuracy gain on ImageNet, $+1.1\%$ mAP gain with YOLOF and $+0.1\%$ mAP gain with FCOS.}

\noindent\textbf{Multi-scale entropy}$\quad$ When using multi-scale entropy, both single-scale model and multi-scale model get similar accuracy on ImageNet. The single-scale model uses 2X more parameters than the multi-scale model under the same FLOPs constraint. In terms of mAP, multi-scale model outperforms single-scale model by $+0.3\%$ on COCO with YOLOF and $0.6\%$ on COCO with FCOS. From the last row of Table~\ref{table:search}, the coarse-to-fine mutation further enhances the performance of multi-scale entropy prior, and the overall improvement over ResNet-50 is $+1.1\%$ on ImageNet-1k, $+2.5\%$ on COCO with YOLOF and $+2.0\%$ on COCO with FCOS.

\noindent\textcolor{black}{\textbf{Correlation between entropy and mAP}$\quad$ 
To further study the correlations between model entropy and model mAP, models during the search are trained, and the results are exhibited in Figure~\ref{fig:corre}. 
The right part of Figure~\ref{fig:corre} indicates that the mAP positively correlates with the multi-scale entropy.
The left part of Figure~\ref{fig:corre} reveals that the single-scale entropy cannot represent the mAP well, so multi-scale entropy is necessary for detection tasks.
By analyzing the structures in Appendix~\ref{app:vsp}, the computation of single-scale models is concentrated in the last stage ${\rm C}{5}$, ignoring the ${\rm C}{3}$ and ${\rm C}{4}$ stages, and leading to the worse multi-scale score. Instead, multi-scale models allocate more computation to the previous stages to enhance the expressivity of ${\rm C}{3}$ and ${\rm C}{4}$, which improve the multi-scale score. 
}

\noindent\textcolor{black}{\textbf{Discussion about dataset}$\quad$ 
We agree that the dataset is powerful for ranking the architectures in the training-based methods. 
However, the process of MAE-DET search is performed without data training.
If we replace the Gaussian input directly with target data, the output after one convolution is also random due to the Gaussian initialized weights. On the other hand, the FPN framework has considered data distribution characteristics according to previous works~\cite{fpn,fcos,gfv2}. Thus, the aim of MAE-Det is to provide a better multi-scale feature extractor under the given inference budgets. We believe the zero-shot method combined with target data without training can be a future research direction.
}

\subsection{Transfer to Other Tasks}
\label{sub:transfer}

\noindent\textbf{VOC and Cityscapes}$\quad$ 
To evaluate the transferability of MAE-DET in different datasets, we transfer the FCOS-based MAE-DET-M to VOC and Cityscapes dataset, as shown in the upper half of Table~\ref{table:mask}. The models are fine-tuned after being pre-trained on ImageNet. Comparing to ResNet-50, MAE-DET-M achieves $+4.1\%$ better mAP in VOC, $+1.1\%$ better mAP in Cityscape.

\noindent\textbf{Instance Segmentation}$\quad$ 
\textcolor{black}{The lower half of Table~\ref{table:mask} reports results of Mask R-CNN~\citep{maskrcnn} and SCNet~\citep{scnet} models for the COCO instance segmentation task with 6X training from scratch. Comparing to ResNet-50, MAE-DET-M achieves better AP and mask AP with similar model size and FLOPs on Mask RCNN and SCNet.}

\section{Conclusion}
In this paper, we revisit the Maximum Entropy Principle in zero-shot NAS for object detection. The proposed MAE-DET achieves competitive detection accuracy with search speed orders of magnitude faster than previous training-based NAS methods. While modern object detection backbones involve more complex building blocks and network topologies, the design of MAE-DET is conceptually simple and easy to implement, demonstrating the grace of ``simple is better'' philosophy. Extensive experiments and analyses on various datasets validate its excellent transferability. 




\bibliography{refs}
\bibliographystyle{icml2022}

\newpage
\appendix
\onecolumn

\section{Training Details}\label{app:detail}
\noindent\textbf{Searching Details}$\quad$ 
In MAE-DET, the evolutionary population $N$ is set as 256 while total iterations $T=96000$. 
Residual blocks and bottleneck blocks are utilized as searching space when comparing with ResNet series backbone~\citep{resnet}.
Following the previous designs~\citep{detnas,spnas,spinenet}, MAE-DET is optimized for the budget of FLOPs according to the target networks, i.e., ResNet-50 and ResNet-101.
To balance the computational complexity and large resolution demand, the resolution in search is set as $384\times384$ for MAE-DET.
When starting the search, the initial structure is composed of 5 downsampling stages with small and narrow blocks to meet the reasoning budget. 
In the mutation, whether the coarse-mutation or fine-mutation, the width of the selected block is mutated in a given scale $\{1/1.5, 1/1.25, 1, 1.25, 1.5, 2\}$, while the depth increases or decreases 1 or 2. The kernel size is searched in set $\{3, 5\}$. Note that blocks deeper than 10 will be divided into two blocks equally to enhance diversity.

\noindent\textbf{Dataset and Training Details}$\quad$ 
For object detection, trainval35k with 115K images in the COCO dataset is mainly used for training. 
With the single-scale testing of resolution $1333\times800$, COCO mAP results are reported on the val 2017 for most experiments and the test-dev 2007 for GFLV2 results in Table~\ref{table:sota}.
When training on the COCO dataset, the initial learning rate is set to $0.02$, and decays two times with the ratio of $0.1$ during training. 
SGD is adopted as optimizer with momentum 0.9; weight decay of $10^{-4}$; batch size of 16 (on 8 Nvidia V100 GPUs); patch size of $1333\times800$.

Additionally, multi-scale training and Synchronized Batch Normalization (SyncBN) are adopted to enhance the stability of the scratch training without increasing the complexity of inference.
Training from scratch is used to avoid the gap between ImageNet pre-trained models, to ensure a fair comparison with baselines.
3X learning schedule is applied for the ablation study with a multi-scale range between $[0.8, 1.0]$ (36 epochs, decays at 28 and 33 epochs), and 6X learning schedule for the SOTA comparisons with the range between $[0.6, 1.2]$ (73 epochs, decays at 65 and 71 epochs).
All object detection training is produced under mmdetection~\citep{mmdet} for fair comparisons, and hyper-parameters not mentioned in the paper are always set to default values in mmdection.

For image classification, all models are trained on ImageNet-1K with a batch size of 256 for 120 epochs. When training on ImageNet-1K, we use SGD optimizer with momentum 0.9; cosine learning rate decay~\citep{cosine}; initial learning rate 0.1; weight decay $4\times10^{-5}$.

\section{\textcolor{black}{MAE-DET for Mobile Device}}\label{app:mbv2}

\begin{table}[h]
	\vspace{-0.4cm}
	\caption{\textcolor{black}{MAE-DET-MB and MobileNetV2 on the COCO with the SSDLite head, which are trained from scratch with 600 epochs at resolution 320. FPS on Pixel 2 is benchmarked on the full model with CPU, FP32, batch size 1. MAE-DET-MB-M-SE means inserting SE modules to MAE-DET-MB-M.}}
	\label{table:mb}
	\begin{center}
		\scalebox{0.82}{
			\begin{tabular}{ccc|cccc|c}
				\toprule[1.5pt]
				& FLOPs & Params & \multicolumn{4}{c|}{val2017} & \multicolumn{1}{c}{FPS}\\
				Backbone & {Backbone} & {Backbone} & ${\bf AP_{val}}$& ${\bf AP}_{S}$ & ${\bf AP}_{M}$ & ${\bf AP}_{L}$ & \makecell[c]{on Pixel 2}\\
				\midrule[1.5pt]
				MobileNetV2-0.5 & 217M & 0.7M & 14.7 & 0.8 & 11.0 & 31.2 & 13.5\\
				MobileNetV2-1.0 & 651M & 2.2M & 21.1 & 1.7 & 20.5 & 39.9 &  6.6\\
				\midrule[1.5pt]
				MAE-DET-MB-S & 201M & 0.6M & 15.9 & 0.8 & 12.2 & 31.8 &  13.8\\
				MAE-DET-MB-M & 645M & 2.0M & 22.2 & 2.1 & 21.5 & 42.3 &  6.3\\
				\midrule[1pt]
				MAE-DET-MB-M-SE & 647M & 2.3M & 22.6 & 2.3 & 22.0 & 42.5 &  5.6\\
				\bottomrule[1.5pt]
		\end{tabular}}
	\end{center}
	\vspace{-0.2cm}
\end{table}

\textcolor{black}{For mobile-size object detection, we explore building MAE-DET-MB with MobileNetV2~\citep{mbv2} blocks, using the inverted bottleneck block with expansion ratio of 1/3/6. The weight ratio $\alpha$ is still set as 1:1:6, and other searching settings are the same as the Resnet-like searching.
In Table.~\ref{table:mb}, MAE-DET-MB use less computation and parameters, but outperform MobileNetV2 by 1\% AP with similar inference time on Google Pixel 2 phone. 
Additionally, inserting SE modules to MAE-DET-MB-M can improve the mAP by 0.4\%.}

\section{Object Detection with ImageNet Pre-train Models}\label{app:pretrain}
In the main body of the paper, training from scratch is used to avoid the gap between ImageNet pre-trained models, to ensure a fair comparison with baselines~\citep{scratch}. Since 6X training from scratch consumes 3 times more time than 2X pre-trained training, we use the ImageNet pre-trained model to initialize the MAE-DET-M in various heads, including \textcolor{black}{RetinaNet}, FCOS and GFLV2. 
As present in Table.~\ref{table:sota},~\ref{table:pretrain}, whether using training from scratch or ImageNet pre-training, MAE-DET can outperform ResNet-50 in the three popular FPN-based frameworks by large margins.

\begin{table}[th]
	\caption{Results between Scratch and Pretrain strategy on the COCO with single-scale testing. Training strategy on ImageNet is same as Table.\ref{table:search}.}
	\label{table:pretrain}
	\begin{center}
		\scalebox{0.82}{
			\begin{tabular}{ccc|ccc|cccc}
				\toprule[1.5pt]
				Backbone & \makecell[c]{FLOPs\\Backbone} & \makecell[c]{Params\\Backbone} & Head  & Strategy & Epochs& ${\bf AP_{val}}$& ${\bf AP}_{S}$ & ${\bf AP}_{M}$ & ${\bf AP}_{L}$\\
				\toprule[1.5pt]
				\multirow{3}*{ResNet-50} & \multirow{3}*{83.6G} & \multirow{3}*{23.5M} &
				GFLV2 & Scratch & 73 & 44.7 & 29.1 & 48.1 & 56.6\\
				&  &  & GFLV2 & Pretrain & 24 & 44.0 & 27.1 & 47.8 & 56.1\\
				&  &  & GFLV2 & Pretrain & 24 & 43.9$\dagger$ & - & - & -\\
				\toprule[1pt]
				\multirow{6}*{MAE-DET-M} & \multirow{6}*{89.9G} & \multirow{6}*{25.8M} &
				\textcolor{black}{RetinaNet} & Scratch & 73  & 42.0 & 26.7 & 45.2 & 55.1\\
				&  &  & \textcolor{black}{RetinaNet} & Pretrain & 24 & 42.3 & 25.3 & 46.5 & 56.0\\
				&  &  & FCOS & Scratch & 73  & 44.5 & 28.6 & 48.1 & 56.1\\
				&  &  & FCOS  & Pretrain & 24 & 44.5 & 28.8 & 48.5 & 56.9\\
				&  &  & GFLV2 & Scratch & 73  & 46.8 & 29.9 & 50.4 & 60.0 \\
				&  &  & GFLV2 & Pretrain & 24  & 46.0 & 29.0 & 50.0 & 59.9 \\
				\toprule[1.5pt]
		\end{tabular}}
	\end{center}
	$\dagger$: results in this line are reported in the official github~\citep{gfv2}. 
\end{table}

\section{Multi-scale Entropy Prior Ablation Study}\label{app:weight}
\begin{table}[th]
	\caption{Results between different ratios of weights $\boldsymbol alpha$ in MSEP on COCO.}
	\label{table:weight}
	\begin{center}
		\scalebox{0.8}{
			\begin{tabular}{cc|cc|ccc|cccc}
				\toprule[1pt]
				Backbone & {$\boldsymbol \alpha_3$:$\boldsymbol\alpha_4$:$\boldsymbol\alpha_5$} & \makecell[c]{FLOPs} & \makecell[c]{Params} & ${\bf AP_{val}}$ & ${\bf AP}_{50}$ & ${\bf AP}_{75}$ & ${\bf  AP}_S$ & ${\bf AP}_M$ & ${\bf AP}_L$ & \\
				\toprule[1pt]
				ResNet-50 & None & 83.6G & 23.5M & 38.0 & 55.2 & 41.0 & 23.2 & 40.8 & 47.6 &  \\
				\toprule[1pt]
				MAE-DET & 1:1:1 & 84.4G & 11.5M & 37.4 & 54.6 & 40.0 & 23.6 & 39.8 & 46.6 &  \\
				MAE-DET & 1:1:2 & 84.8G & 13.4M & 37.8 & 54.9 & 40.5 & 23.2 & 40.0 & 47.8 &  \\
				MAE-DET & 1:1:4 & 85.9G & 17.2M & 38.6 & 56.0 & 41.4 & 23.4 & 41.3 & 48.6 &  \\
				MAE-DET & 1:1:6 & 88.7G & 29.4M & \textbf{39.4} & \textbf{57.3} & \textbf{42.1} & \textbf{23.7} & 42.3 & \textbf{50.0} &  \\
				MAE-DET & 1:1:8 & 89.9G & 31.7M & \textbf{39.4} & 57.2 & 42.0 & \textbf{23.7} & \textbf{42.5} & 49.5  & \\
				\toprule[1pt]
				MAE-DET & 1:4:1 & 86.3G & 10.9M & 35.7 & 52.6 & 38.3 & 22.2 & 38.1 & 44.9 &  \\
				MAE-DET & 4:1:1 & 86.1G & 11.1M & 33.9 & 50.2 & 36.7 & 20.4 & 36.1 & 43.4 &  \\
				\toprule[1pt]
		\end{tabular}}
	\end{center}
\end{table}

\begin{figure}[h]
	\centering
	\vspace{-0.2cm}
	\includegraphics[scale=0.45]{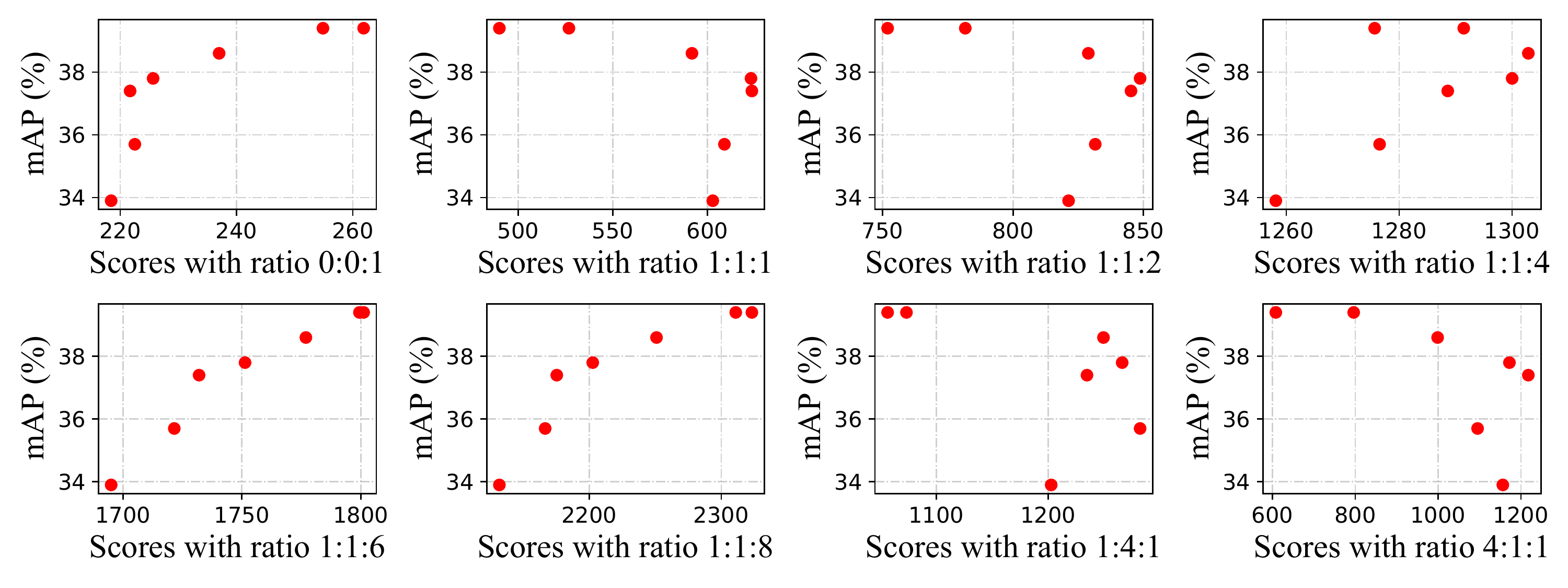}
	\vspace{-0.2cm}
	\caption{\textcolor{black}{mAP vs. scores. All models are from Table~\ref{table:weight} and the scores are computed with different weight ratios. When the ratio is equal to 1:1:6, the correlation between mAP and score is well fitted.}}
	\label{fig:alpha}
	\vspace{-0.2cm}
\end{figure}

\textcolor{black}{In Table~\ref{table:weight}, we tune the different arrangements of multi-scale weights in a wide range. Seven multi-scale weight ratios are used to search different models, and all models are trained on the COCO dataset with FCOS and 3X learning schedule. 
Table~\ref{table:weight} shows that if the same weights are arranged to ${\rm C}{3}$-${\rm C}{5}$, the performance of MAE-DET on COCO is worse than ResNet-50. 
Considering the importance of ${\rm C}{5}$ (discussed in Section~\ref{sub:msep}), we increase the weight of ${\rm C}{5}$, and MAE-DET's performance continues to improve. 
To further explore the correlations between mAP and scores, we use the seven weight ratios to calculate the different scores of each model, along with the single-scale weight ratio of 0:0:1. The correlations between mAP and different scores are plotted in Figure~\ref{fig:alpha}.
Taking the results in Table~\ref{table:weight} and Figure~\ref{fig:alpha}, we confirm the ratio of 1:1:6 may be good enough for the current FPN structure.}

\section{Comparison with Zero-Shot Proxies for Image Classification}\label{app:zero-shot}

\begin{table}[th]
	\caption{Different zero-shot proxies on COCO with FCOS. All methods use the same search space, FLOPs budget, searching strategy and training schedule.}
	\label{table:zsm}
	\begin{center}
		\scalebox{0.8}{
			\begin{tabular}{c|cc|cccc}
				\toprule[1pt]
				Proxy & \makecell[c]{FLOPs\\Backbone} & \makecell[c]{Params\\Backbone}
				& ${\bf AP_{val}}$ & ${\bf AP}_{S}$ & ${\bf AP}_{M}$ & ${\bf AP}_{L}$\\
				\midrule[1pt]
				ResNet-50 &  84G & 23.5M & 38.0 & 23.2 & 40.8 & 47.6\\
				\midrule[1pt]
				SyncFlow & 90G & 67.4M & 35.6 & 21.8 & 38.1 & 44.8\\
				NASWOT & 88G & 28.1M & 36.7 & 23.1 & 38.8 & 45.9\\
				Zen-NAS & 91G & 67.9M & 38.1 & 23.2 & 40.5 & 48.1\\
				MAE-DET & 89G & 25.8M & \textbf{39.4} & 23.7 & 42.3 & 50.0 \\
				\bottomrule[1pt]
		\end{tabular}}
	\end{center}
    \vspace{-0.4cm}
\end{table}
We compare MAE-DET with architectures designed by zero-shot proxies for image classification in previous works, including SyncFlow~\citep{syn}, NASWOT~\citep{naswot}, ZenNAS~\citep{zennas}. For a fair comparison, all methods use the same search space, FLOPs budget $91\,G$, searching strategy and training schedule. All searched backbones are trained on COCO with the FCOS head and 3X training from scratch. The results are reported in Table.~\ref{table:zsm}. 

Among these methods, SyncFlow and NASWOT perform worse than ResNet-50 on COCO albeit they show competitive performance in image classification tasks. Zen-NAS achieves competitive performance over ResNet-50. The MAE-DET outperforms Zen-NAS by $+1.3\%$ mAP with slightly fewer FLOPs and nearly one third of parameters.

\section{Random search comparisons}\label{app:random}

\begin{figure}[h]
	\centering
	\subfigure{\includegraphics[scale=0.5]{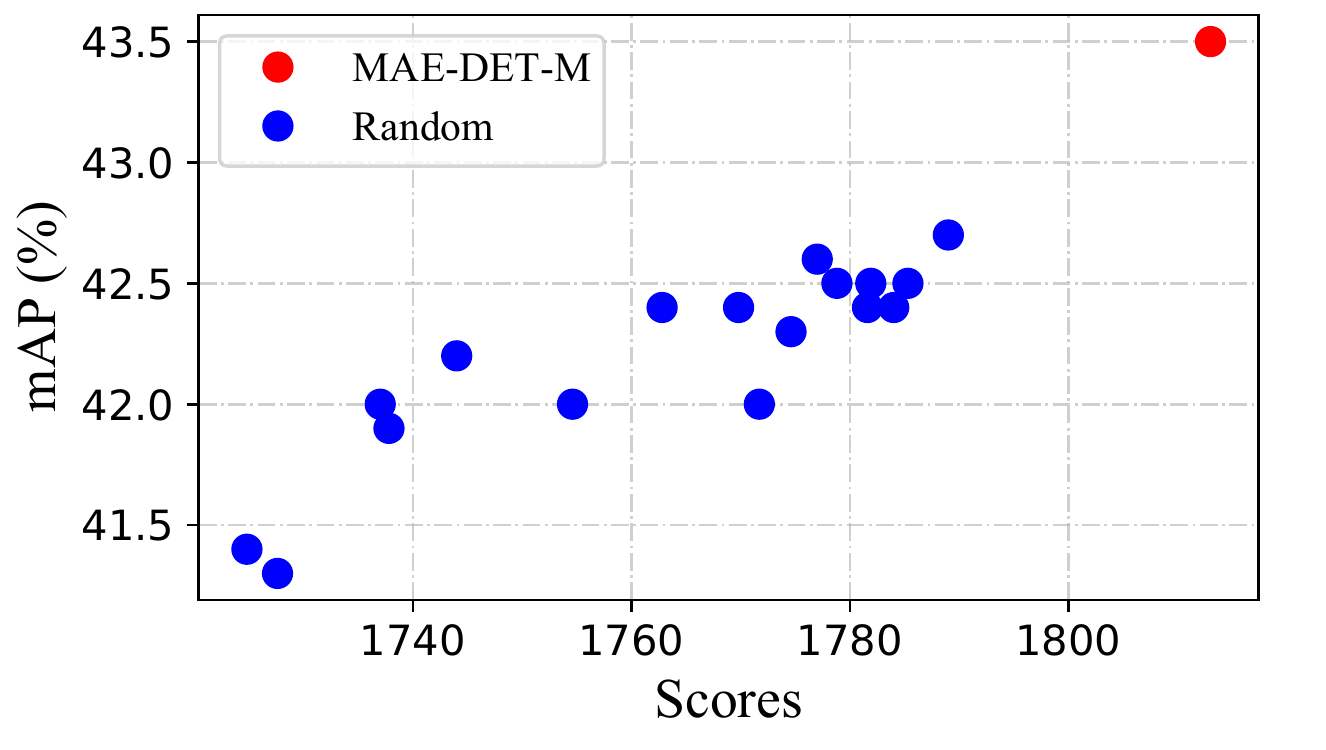}}
	\subfigure{\includegraphics[scale=0.5]{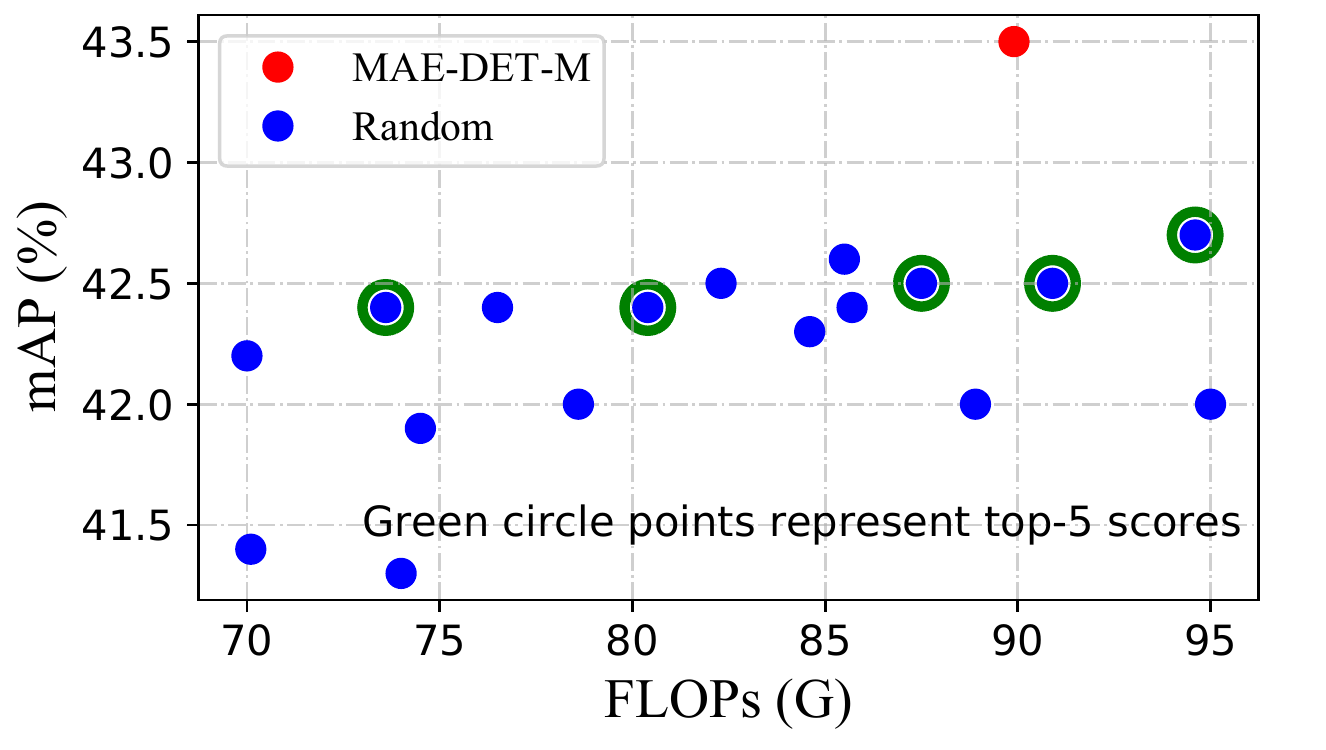}}
	\caption{Correlations between random sample models.}
	\label{fig:sample}	
\end{figure}

Using the same search space as MAE-DET-M, we randomly sample models within 70-95G FLOPs and 20-30M parameters without MAE score. Due to the high cost of sampling, 17 random models are currently sampled and trained on GFLV2 with a 3X learning schedule. The correlations are shown in Fig.~\ref{fig:sample}. Our MAE-DET model has better performance than the random searched models.

\vspace{-8pt}

\section{Visualization of Search process}\label{app:vsp}

\begin{figure}[h]
	\centering
	\subfigure[Iterations vs. \#Layers]{\includegraphics[scale=0.22]{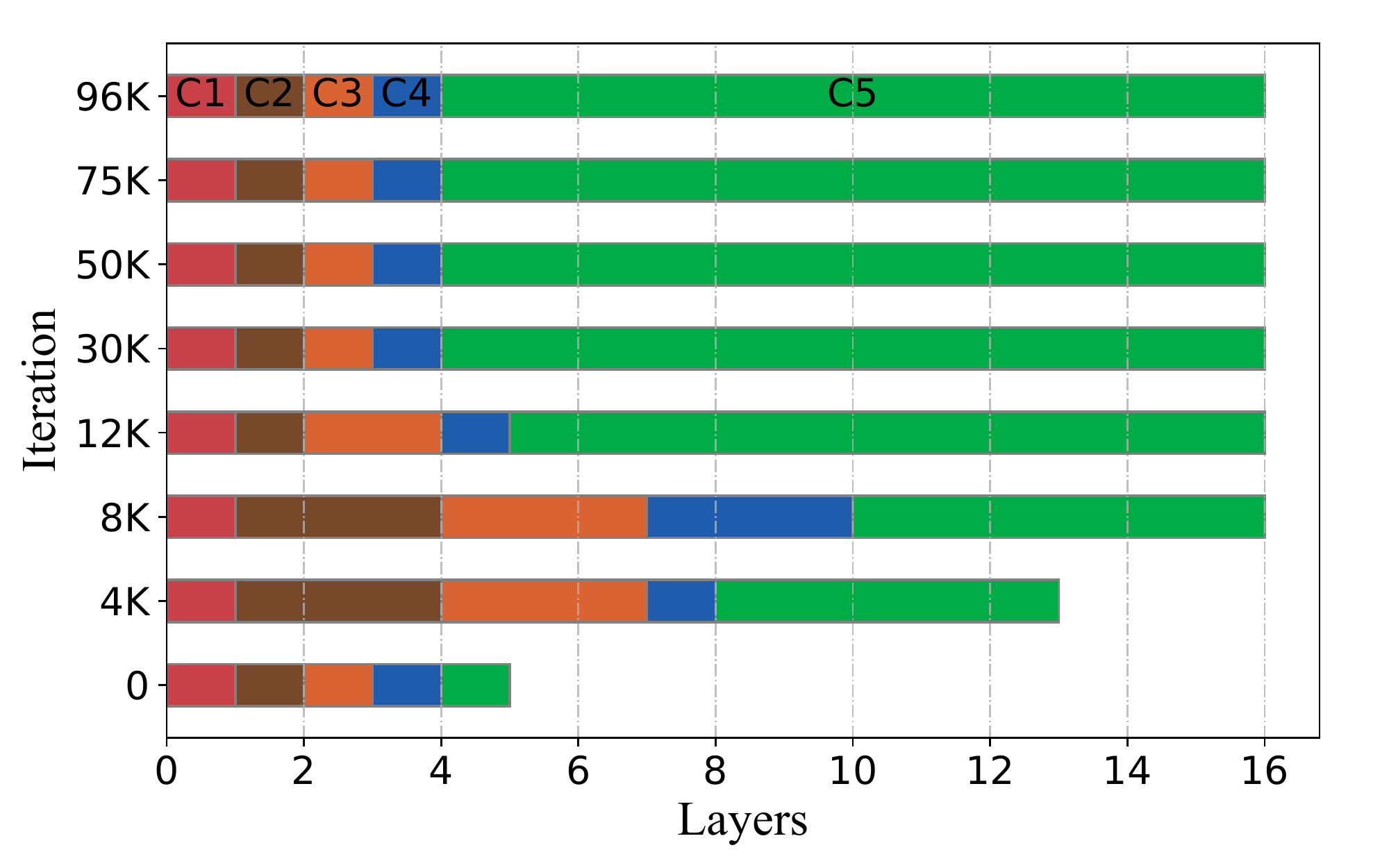} \label{sfig:l3}}
	\subfigure[Iterations vs. FLOPs]{\includegraphics[scale=0.22]{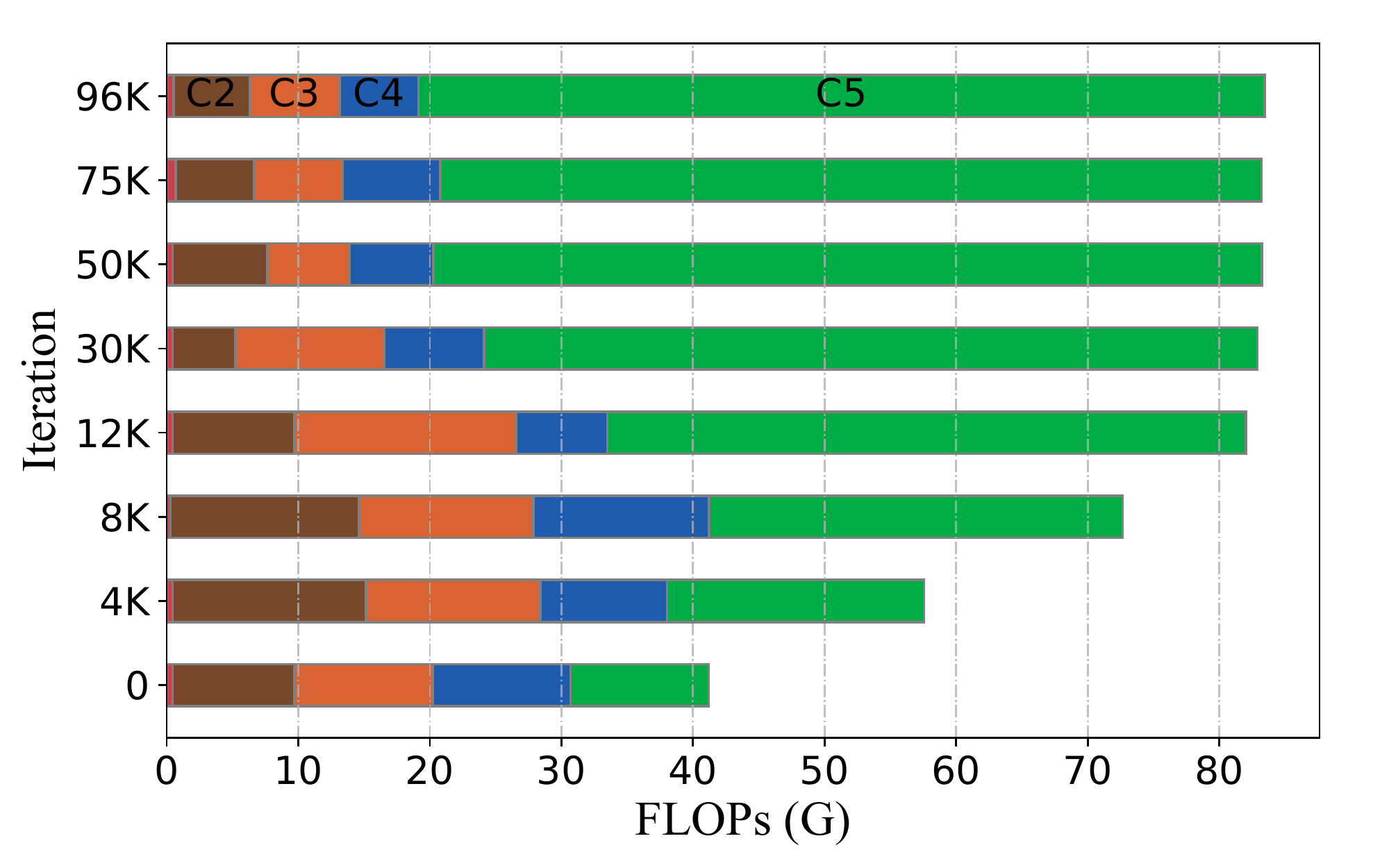} \label{sfig:f3}}
	\subfigure[IterationsP vs. Parameters]{\includegraphics[scale=0.22]{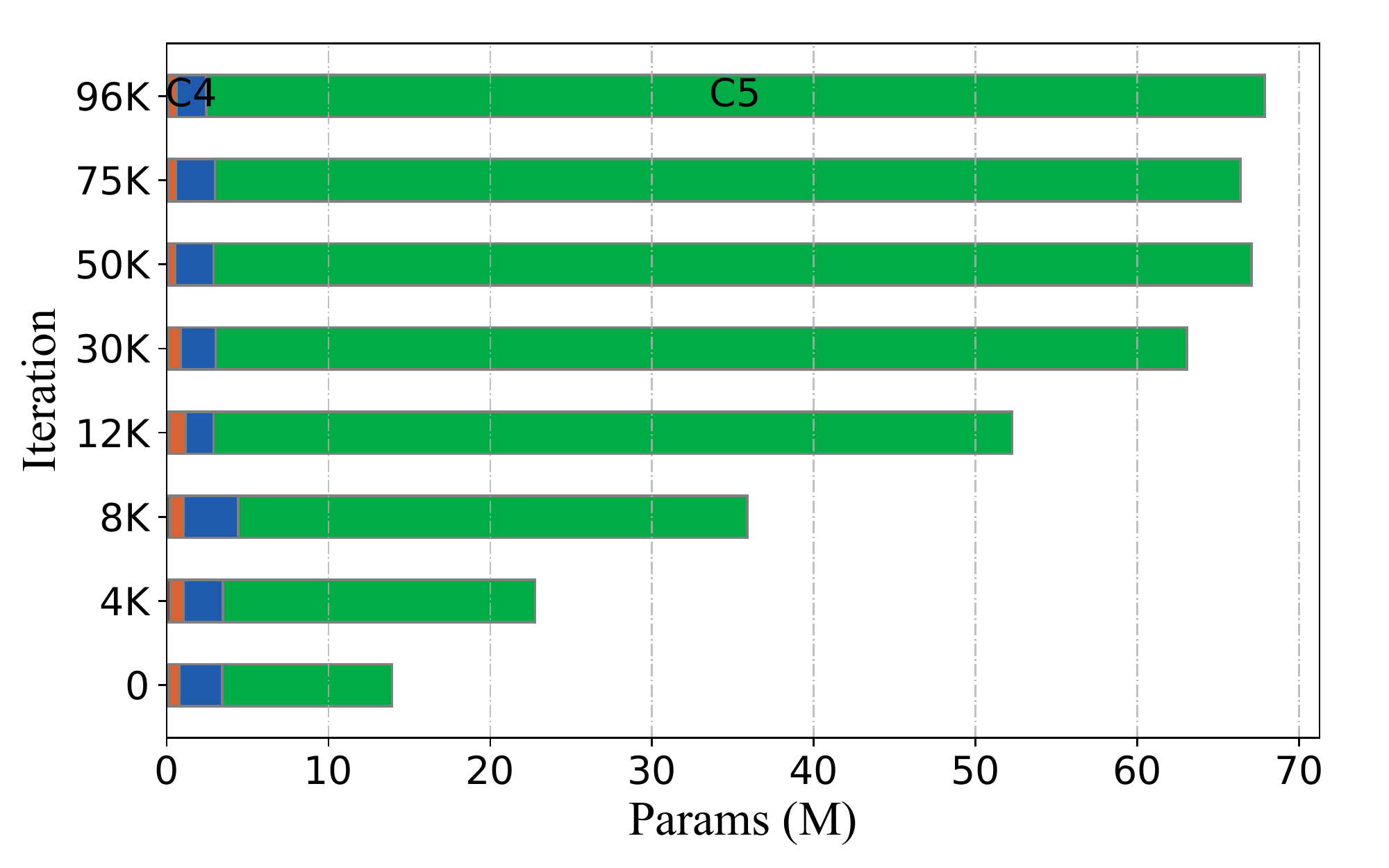} \label{sfig:p3}}
	\caption{Visualization of single-scale entropy searching process. \#layer is the number of each block of different levels.}
	\label{fig:sse}
\end{figure}
\begin{figure}[h]
	\centering
	\subfigure[Iterations vs. \#Layers]{\includegraphics[scale=0.22]{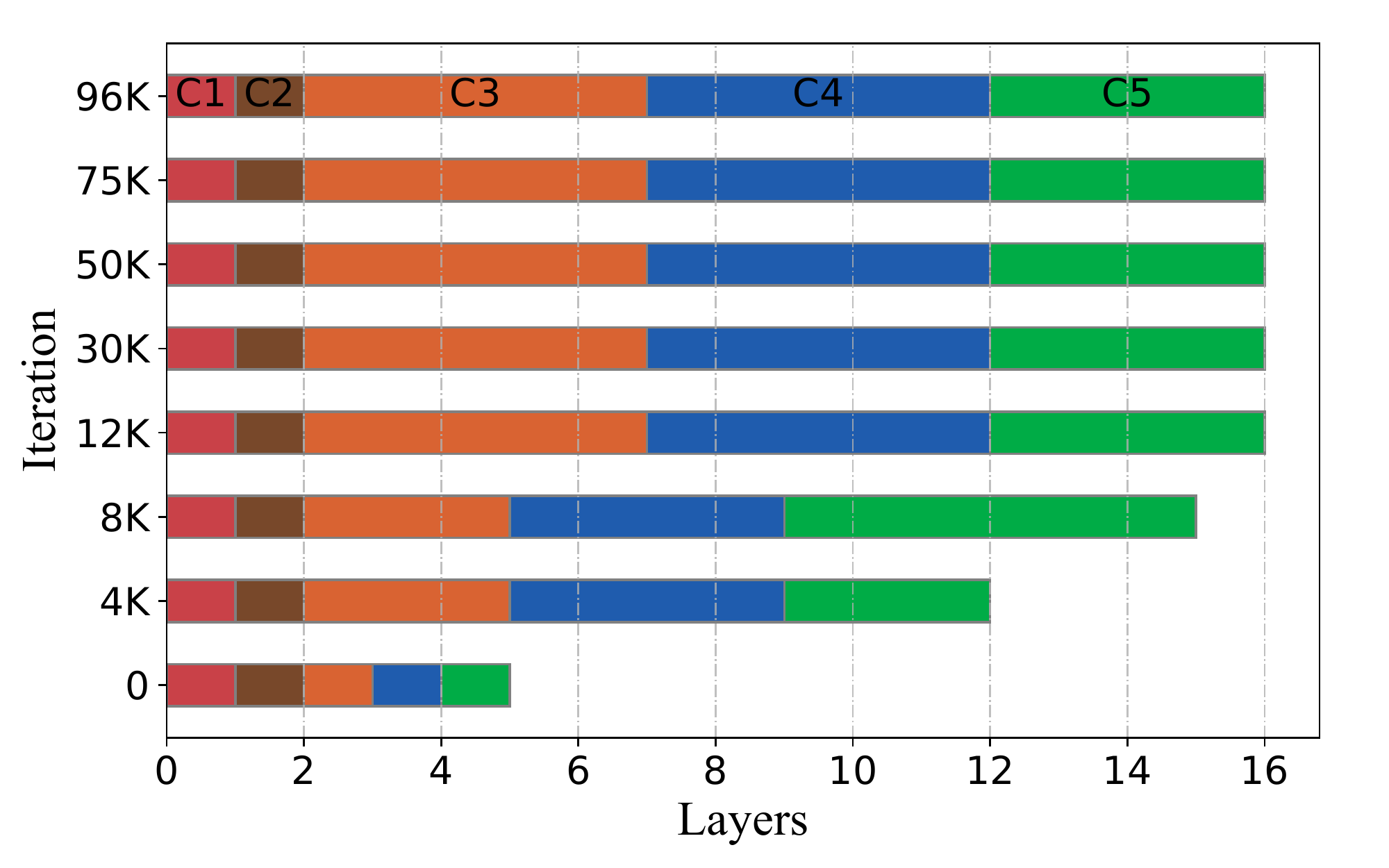} \label{sfig:l2}}
	\subfigure[Iterations vs. FLOPs]{\includegraphics[scale=0.22]{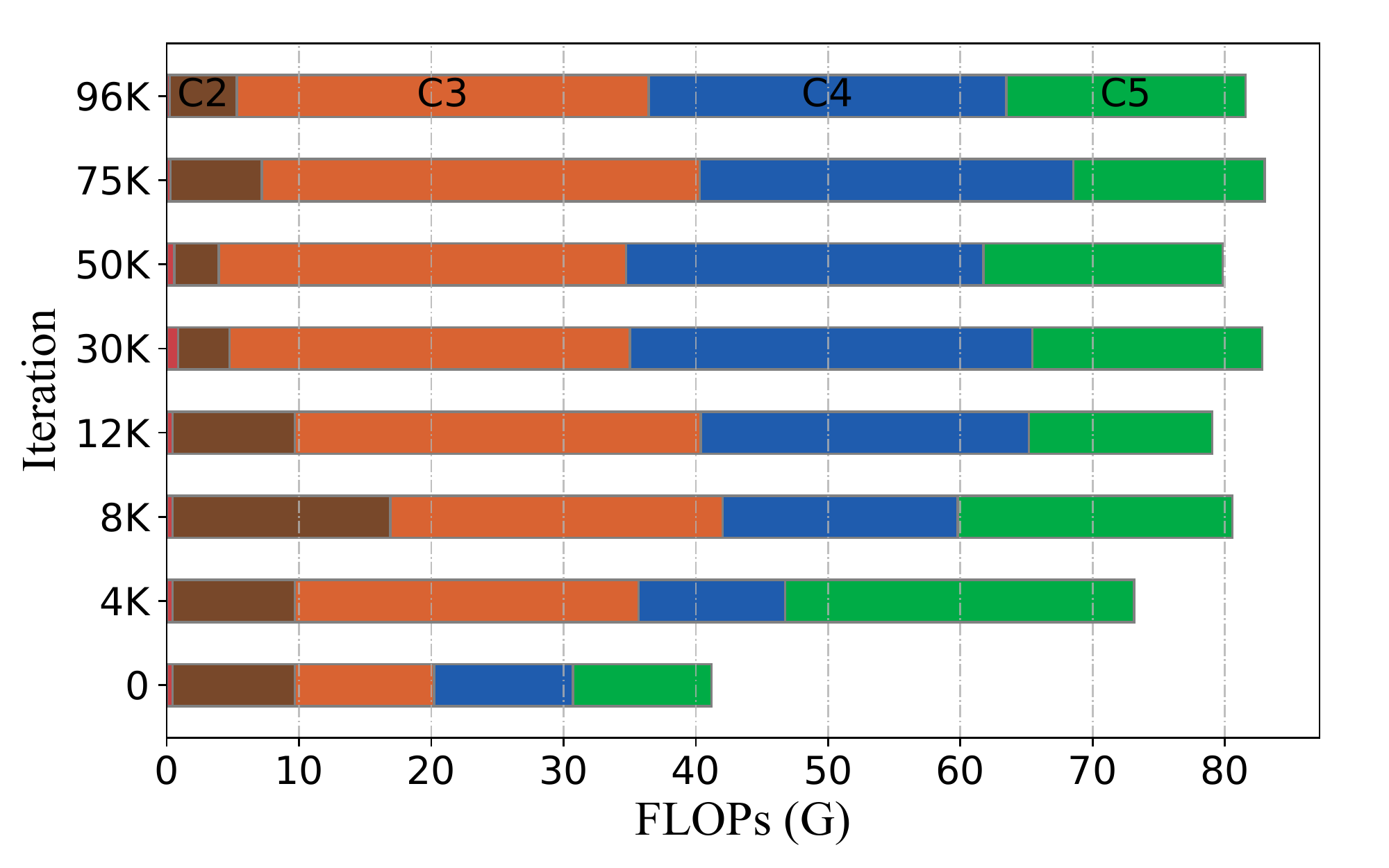} \label{sfig:f2}}
	\subfigure[IterationsP vs. Parameters]{\includegraphics[scale=0.22]{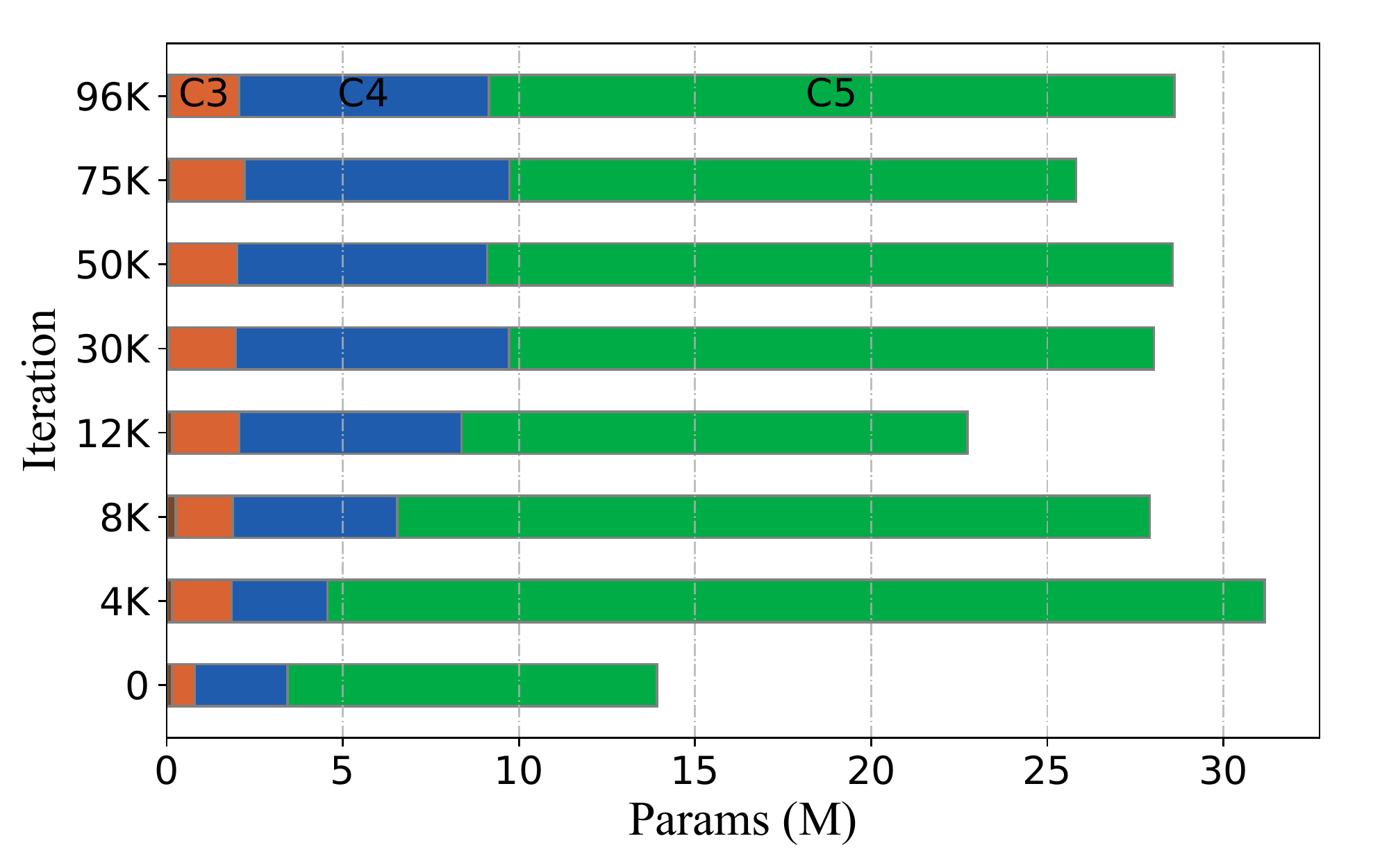} \label{sfig:p2}}
	\caption{Visualization of multi-scale entropy searching process (Coarse). \#layer is the number of each block of different levels.}
	\label{fig:mse-coarse}
\end{figure}
\begin{figure}[!h]
	\centering
	\subfigure[Iterations vs. \#Layers]{\includegraphics[scale=0.22]{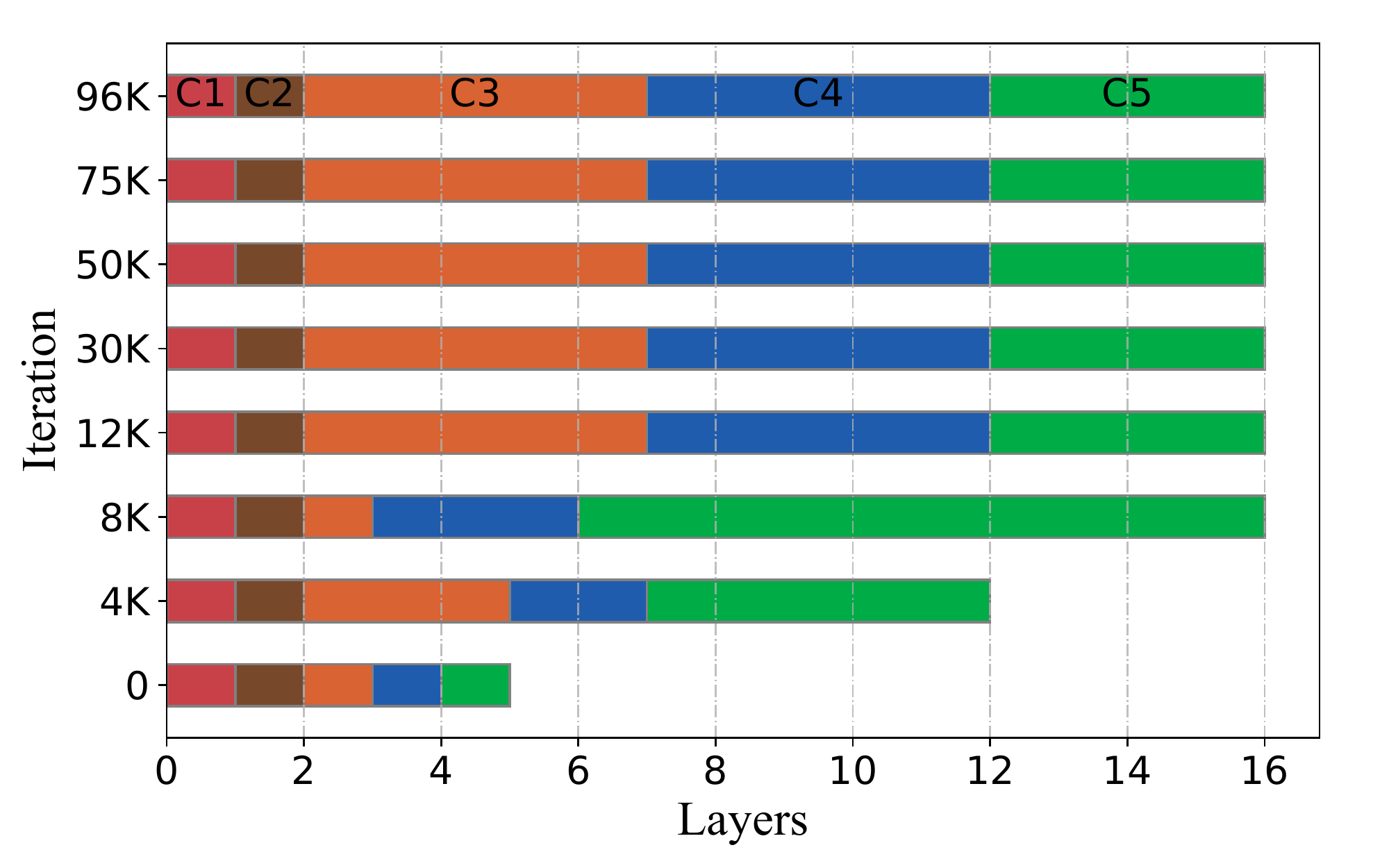} \label{sfig:l1}}
	\subfigure[Iterations vs. FLOPs]{\includegraphics[scale=0.22]{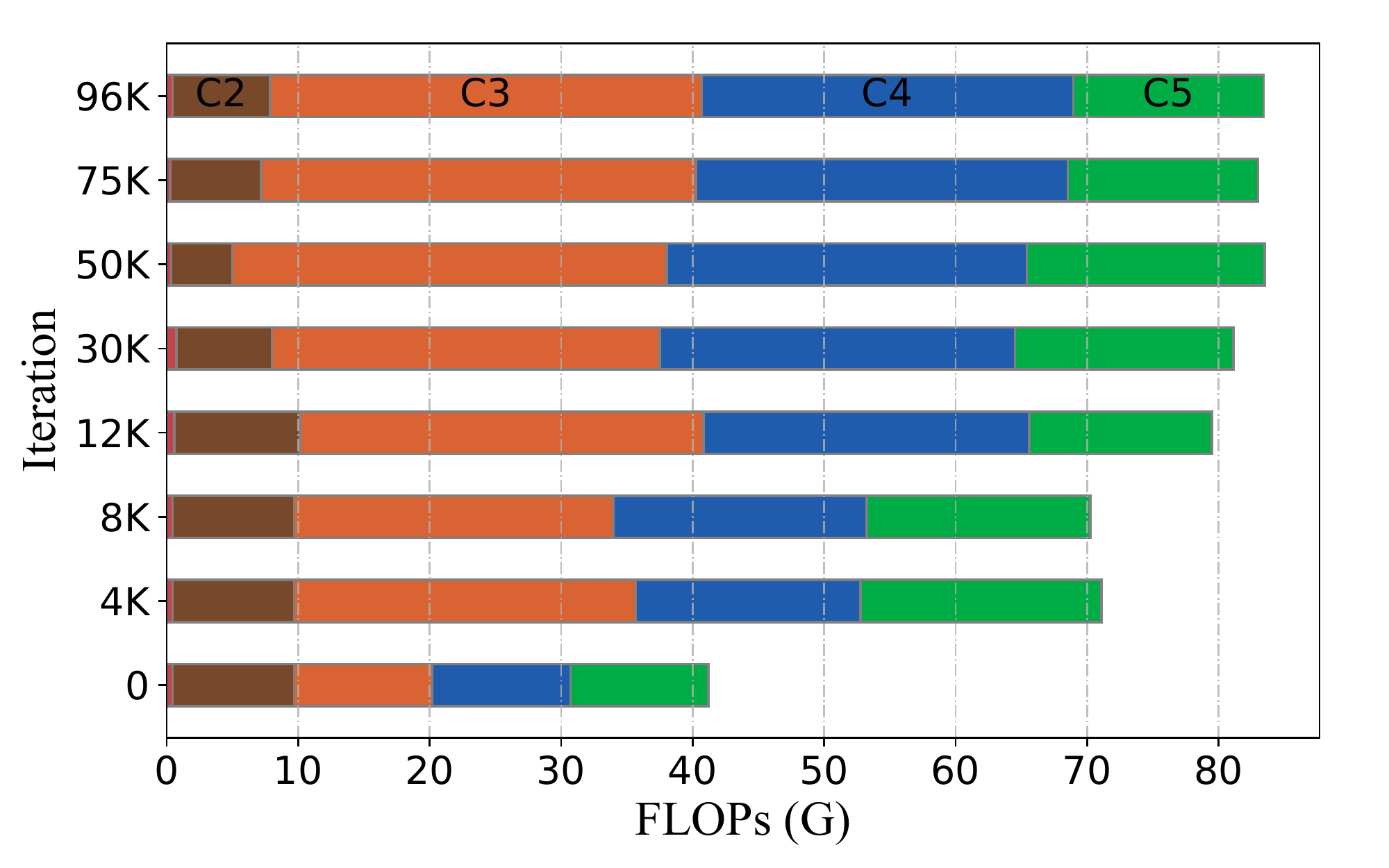} \label{sfig:f1}}
	\subfigure[IterationsP vs. Parameters]{\includegraphics[scale=0.22]{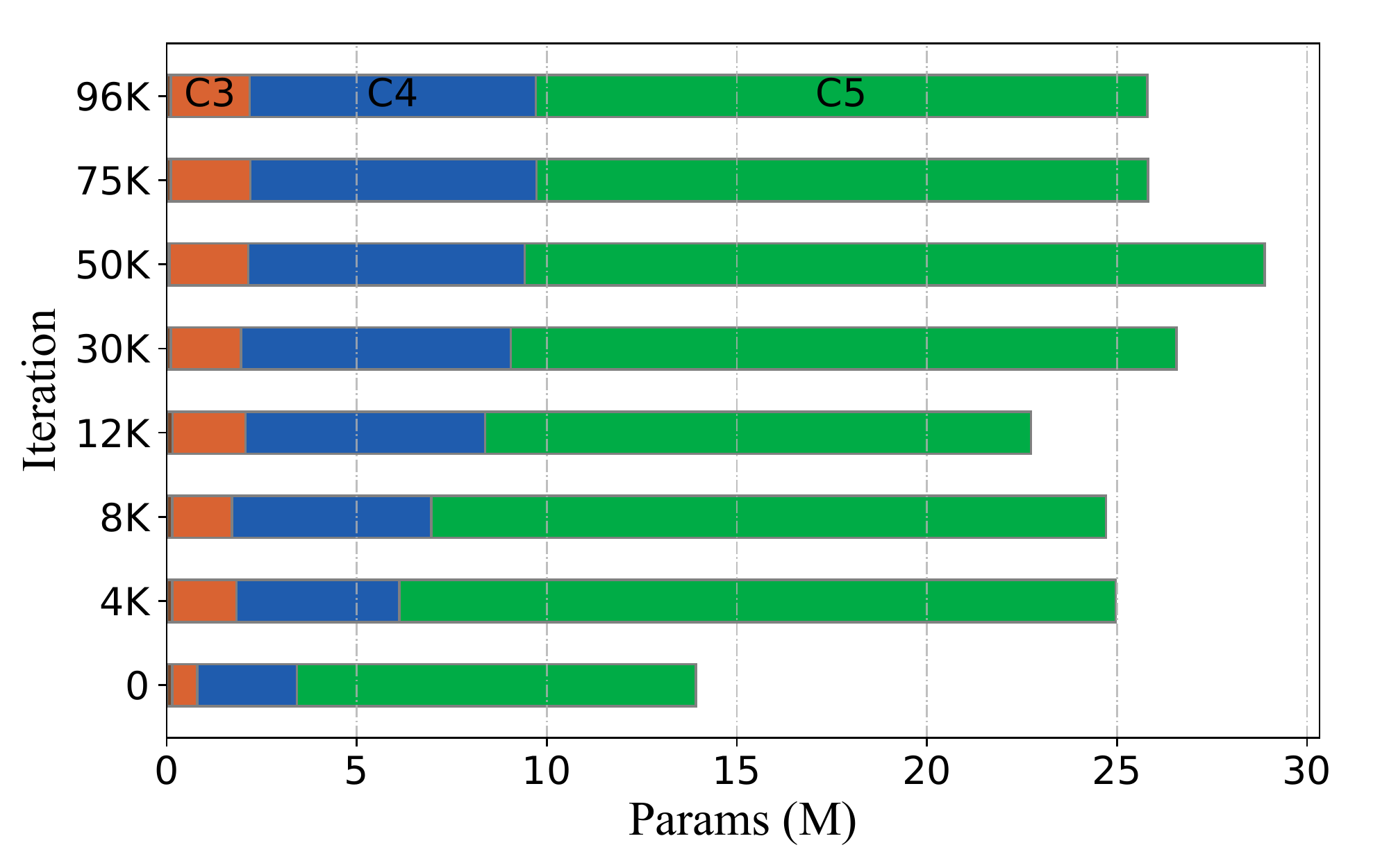} \label{sfig:p1}}
	\caption{Visualization of single-scale entropy searching process (Coarse-to-Fine). \#layer is the number of each block of different levels.}
	\label{fig:mse-fine}
\end{figure}

The visualizations of the searching process on different entropies are shown in Figure~\ref{fig:sse},~\ref{fig:mse-coarse},~\ref{fig:mse-fine}. We compare the change of layers, FLOPs and parameters during different iterations. Visualizations prove our assumptions in the main body of the paper.

\section{Detail Structure of MAE-DETs}\label{app:struc}
We list detail structure in Table~\ref{table:sota}, \ref{table:search}.

The 'block' column is the block type. 'Conv' is the standard convolution layer followed by BN and RELU. 'ResBlock' is the residual bottleneck block used in ResNet-50 and is stacked by two Blocks in our design. 'kernel' is the kernel size of kxk convolution layer in each block. 'in', 'out' and 'bottleneck' are numbers of input channels, output channels and bottleneck channels respectively. 'stride' is the stride of current block. '\# layers' is the number of duplication of current block type.

\begin{table}[h]
	\caption{Architecture of single-scale entropy with coarse mutation in Table.~\ref{table:search}}
	\label{tab:sse}
	\begin{center}
		\scalebox{0.8}{
		\begin{tabular}{clclclclclclclclc}
			\toprule 
			block & kernel & in & out & stride & bottleneck & \# layers& level\tabularnewline
			\midrule
			\midrule
			Conv & 3 & 3 & 96 & 2 & - & 1 & C1\tabularnewline
			\midrule
			ResBlock & 5 & 96 & 208 & 2 & 32 & 2 & C2\tabularnewline
			\midrule
			ResBlock & 5 & 208 & 560 & 2 & 56 & 1 & C3\tabularnewline
			\midrule
			ResBlock & 5 & 560 & 1264 & 2 & 112 & 2 & C4\tabularnewline
			\midrule
			ResBlock & 5 & 1264 & 1712 & 2 & 224 & 3 & C5\tabularnewline
			\midrule
			ResBlock & 5 & 1712 & 2048 & 1 & 224 & 3 & C5\tabularnewline
			\midrule
			ResBlock & 5 & 2048 & 2048 & 1 & 256 & 4 & C5\tabularnewline
			\bottomrule
		\end{tabular}}
	\end{center}
\end{table}

\begin{table}[h]
	\caption{Architecture of multi-scale entropy with coarse mutation in Table.~\ref{table:search}}
	\label{tab:mse}
	\begin{center}
		\scalebox{0.8}{
		\begin{tabular}{clclclclclclclclc}
			\toprule 
			block & kernel & in & out & stride & bottleneck & \# layers& level\tabularnewline
			\midrule
			\midrule
			Conv & 3 & 3 & 32 & 2 & - & 1 & C1\tabularnewline
			\midrule
			ResBlock & 5 & 32 & 128 & 2 & 24 & 1 & C2\tabularnewline
			\midrule
			ResBlock & 5 & 128 & 512 & 2 & 72 & 5 & C3\tabularnewline
			\midrule
			ResBlock & 5 & 512 & 1632 & 2 & 112 & 5 & C4\tabularnewline
			\midrule
			ResBlock & 5 & 1632 & 2048 & 2 & 216 & 4 & C5\tabularnewline
			\bottomrule
		\end{tabular}}
	\end{center}
\end{table}

\begin{table}[h]
	\caption{Architecture of multi-scale entropy with coarse-to-fine mutation in Table.~\ref{table:search} / MAE-DET-M architecture in Table.~\ref{table:sota}}
	\label{tab:MAE-DET-m}
	\begin{center}
		\scalebox{0.8}{
		\begin{tabular}{clclclclclclclclc}
			\toprule 
			block & kernel & in & out & stride & bottleneck & \# layers& level\tabularnewline
			\midrule
			\midrule
			Conv & 3 & 3 & 64 & 2 & - & 1 & C1\tabularnewline
			\midrule
			ResBlock & 3 & 64 & 120 & 2 & 64 & 1 & C2\tabularnewline
			\midrule
			ResBlock & 5 & 120 & 512 & 2 & 72 & 5 & C3\tabularnewline
			\midrule
			ResBlock & 5 & 512 & 1632 & 2 & 112 & 5 & C4\tabularnewline
			\midrule
			ResBlock & 5 & 1632 & 2048 & 2 & 184 & 4 & C5\tabularnewline
			\bottomrule
		\end{tabular}}
	\end{center}
\end{table}

\newpage
\begin{table}[!h]
	\caption{MAE-DET-S in Table.~\ref{table:sota}}
	\label{tab:MAE-DET-s}
	\begin{center}
		\scalebox{0.8}{
		\begin{tabular}{clclclclclclclclc}
			\toprule 
			block & kernel & in & out & stride & bottleneck & \# layers& level\tabularnewline
			\midrule
			\midrule
			Conv & 3 & 3 & 32 & 2 & - & 1 & C1\tabularnewline
			\midrule
			ResBlock & 5 & 32 & 48 & 2 & 32 & 1 & C2\tabularnewline
			\midrule
			ResBlock & 3 & 48 & 272 & 2 & 120 & 2 & C3\tabularnewline
			\midrule
			ResBlock & 5 & 272 & 1024 & 2 & 80 & 5 & C4\tabularnewline
			\midrule
			ResBlock & 3 & 1024 & 2048 & 2 & 240 & 5 & C5\tabularnewline
			\bottomrule
		\end{tabular}}
	\end{center}
\end{table}

\begin{table}[!h]
	\caption{MAE-DET-L in Table.~\ref{table:sota}}
	\label{tab:MAE-DET-l}
	\begin{center}
		\scalebox{0.8}{
		\begin{tabular}{clclclclclclclclc}
			\toprule 
			block & kernel & in & out & stride & bottleneck & \# layers& level\tabularnewline
			\midrule
			\midrule
			Conv & 3 & 3 & 80 & 2 & - & 1 & C1\tabularnewline
			\midrule
			ResBlock & 3 & 80 & 144 & 2 & 80 & 1 & C2\tabularnewline
			\midrule
			ResBlock & 5 & 144 & 608 & 2 & 88 & 6 & C3\tabularnewline
			\midrule
			ResBlock & 5 & 608 & 1912 & 2 & 136 & 6 & C4\tabularnewline
			\midrule
			ResBlock & 5 & 1912 & 2400 & 2 & 220 & 5 & C5\tabularnewline
			\bottomrule
		\end{tabular}}
	\end{center}
\end{table}

\begin{table}[!h]
	\textcolor{black}{\caption{Initial structure in the search}}
	\label{tab:iinitial}
	\begin{center}
		\scalebox{0.8}{
			\begin{tabular}{clclclclclclclclc}
				\toprule 
				block & kernel & in & out & stride & bottleneck & \# layers& level\tabularnewline
				\midrule
				\midrule
				Conv & 3 & 3 & 64 & 2 & - & 1 & C1\tabularnewline
				\midrule
				ResBlock & 3 & 64 & 256 & 2 & 64 & 1 & C2\tabularnewline
				\midrule
				ResBlock & 3 & 256 & 512 & 2 & 128 & 1 & C3\tabularnewline
				\midrule
				ResBlock & 3 & 512 & 1024 & 2 & 256 & 1 & C4\tabularnewline
				\midrule
				ResBlock & 3 & 1024 & 2048 & 2 & 512 & 1 & C5\tabularnewline
				\bottomrule
		\end{tabular}}
	\end{center}
\end{table}

\end{document}